\documentclass[11pt]{article}

\usepackage[final]{acl}

\usepackage{booktabs}
\usepackage{times}
\usepackage{latexsym}
\usepackage{algpseudocode}
\usepackage{caption}
\usepackage{multirow}
\usepackage{multicol}
\usepackage{arydshln}
\usepackage{amsmath}
\usepackage{amsfonts}
\usepackage{microtype}
\usepackage{ragged2e}
\usepackage{array,tabularx,caption,boldline}
\usepackage{inconsolata}
\usepackage{algorithm}
\usepackage{tabularray}
\usepackage{makecell}
\usepackage{makebox}
\usepackage{enumitem}
\usepackage[table,xcdraw]{xcolor}
\usepackage{tcolorbox}
\definecolor{gray}{RGB}{240,240,240}
\definecolor{darkred}{RGB}{255,128,128}
\definecolor{lightred}{RGB}{255,217,217}
\definecolor{sentlevel}{RGB}{211,255,250}
\definecolor{ctxlevel}{RGB}{255,232,205}
\definecolor{sentctxlevel_raw}{RGB}{255,204,153}
\definecolor{sentctxlevel}{RGB}{255,232,255}
\definecolor{our_raw}{RGB}{234,235,139}
\definecolor{our}{RGB}{217,255,178}

\newcommand{\ignore}[1]{}

\usepackage[T1]{fontenc}

\usepackage[utf8]{inputenc}

\usepackage{microtype}

\usepackage{inconsolata}

\usepackage{graphicx}
\usepackage{tikz}
\usepackage{nicematrix}
\usepackage[normalem]{ulem}
\NiceMatrixOptions
       {
         custom-line =
           {
             letter = ; ,
             command = hdashedline ,
             ccommand = cdashedline,
             tikz = dashed
    } 
}
\title{Cross-Preference Learning for Sentence-Level and Context-Aware Machine Translation}

\author{
 \textbf{Ying Li\textsuperscript{1}},
 \textbf{Xinglin Lyu\textsuperscript{2}},
 \textbf{Junhui Li\textsuperscript{1,3}\thanks{Corresponding author: Junhui Li}},
 \textbf{Jinlong Yang\textsuperscript{4}},
\\
 \textbf{Hengchao Shang\textsuperscript{4}},
 \textbf{Min Zhang\textsuperscript{4}},
 \textbf{Shimin Tao\textsuperscript{4}},
 \textbf{Daimeng Wei\textsuperscript{4}},
\\
 \textsuperscript{1}School of Computer Science and Technology, Soochow University, Suzhou, China
 \\
 \textsuperscript{2}School of Computer Science and Artificial Intelligence, Zhengzhou University, Zhengzhou, China
 \\
 \textsuperscript{3}Jiangsu Key Lab of Language Computing, China
 \\
 \textsuperscript{4}Huawei Translation Services Center, Beijing, China
\\
\texttt{yli2@stu.suda.edu.cn, xllyu@zzu.edu.cn, jhli@suda.edu.cn}
\\
\texttt{\{yangjinlong7,shanghengchao,zhangmin186,taoshimin,weidaimeng\}@huawei.com}
}

\begin{document}
\maketitle
\begin{abstract}
Context-aware machine translation (MT) leverages document-level information, yet it does not consistently outperform sentence-level MT, as contextual signals are unevenly beneficial across sentences. Existing training objectives do not explicitly model this variability, limiting a model's ability to adaptively exploit context. In this paper, we propose Cross-Preference Learning (CPL), a preference-based training framework that explicitly captures the complementary benefits of sentence-level and context-aware MT. CPL achieves this by integrating both intra- and cross-condition preferences into the preference optimization objective, providing explicit supervision to exploit informative context while remaining robust to uninformative context. We validate the proposed approach on several public context-aware MT tasks using multiple models, including Qwen3-4B, Qwen3-8B, and Llama-3-8B-Instruct. Experimental results demonstrate consistent improvements in translation quality and robustness across both input conditions, achieved without any architectural modifications. 
\end{abstract}

\section{Introduction}\label{sec:introduction}

Modern machine translation (MT) systems are increasingly expected to operate under multiple input conditions. In some scenarios, only a single source sentence is available, while in others, additional inter-sentence context can be provided to improve translation quality (e.g., document-level or context-aware MT)~\cite{tiedemann-2017-neural,maruf-2021-survey,appicharla-2025-beyond}. These two settings, sentence-level and context-aware translation, share the same output space but are typically handled by separate models and optimized independently. 

\begin{figure}[!t]
\centering
\includegraphics[width=1.0\columnwidth, trim={0cm 0.3cm 0cm 0cm}]{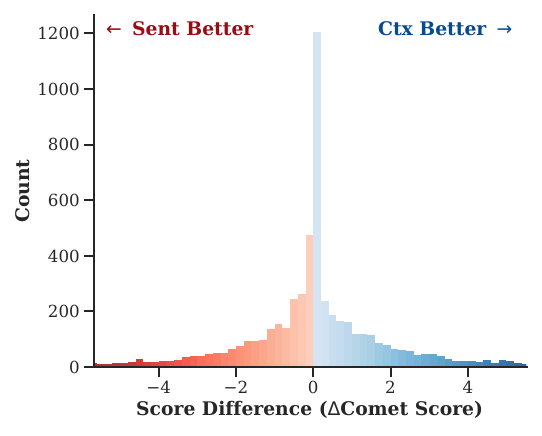}
\caption{Distribution of COMET differences between sentence-level (Sent) and context-aware (Ctx) English–German translations produced by Qwen3-8B.} 
\label{fig:distribution}
\end{figure}
A key observation underlying our work is that context-aware translation does not consistently outperform sentence-level translation. As illustrated in Figure~\ref{fig:distribution}, context-aware models greatly outperform sentence-level models for a subset of sentences, but also slightly underperform, or even greatly underperform for a non-negligible portion of the data, while performing on par in many cases. This distribution highlights that the two systems often produce diverse yet complementary translations rather than one uniformly dominating the other~\cite{voita-2018-context,dong-2025-two}. 
Consistent with this observation, Table~\ref{tab:oracle} shows that the two systems achieve comparable overall BLEU and COMET scores, while the oracle results indicate substantial headroom when the better output is selected on a per-sentence basis. These findings highlight a key limitation of existing approaches: current MT systems lack an explicit mechanism to decide \emph{when} to rely on context. Instead, models are trained under a fixed input condition, implicitly assuming that more context is always beneficial. This leads to suboptimal behavior when context is noisy, redundant, or even misleading.\footnote{Additional analyses of context sensitivity and reranking performance are provided in Appendix~\ref{apx:distribution_analysis} and Appendix~\ref{apx:reranking}.}

Inspired by recent advances in preference optimization, we cast the preference for using inter-sentence context in translation as a preference optimization problem, addressed via preference learning. However, existing methods such as direct preference optimization (DPO)~\cite{rafailov-2023-dpo}, contrastive preference optimization (CPO)~\cite{xu-2024-cpo}, and simple preference optimization (SimPO)~\cite{yu-2024-simpo} focus on single-condition settings, where preference comparisons share the same input configuration. Consequently, these approaches are not designed to account for multiple input conditions sharing the same output space but differ in available information.

\begin{table}[!t]
\centering
\small
\begin{NiceTabular}{l|cc}
\toprule
\bf Model & \bf BLEU & \bf COMET \\
\midrule
Sentence-level   & 31.72 & 86.12 \\
Context-aware & 31.60 & 86.13 \\
\hdashedline
\textit{Oracle} & 33.91 & 87.17 \\
\bottomrule
\end{NiceTabular}
\caption{Overall English-to-German translation quality of sentence-level and context-aware translation. \textit{Oracle} means we always select the better translation from the two systems for each sentence.}
\label{tab:oracle}
\end{table}

To address this limitation, we propose Cross-Preference Learning (CPL) built upon CPO. In addition to the intra-condition preference learning in vanilla CPO, CPL introduces a cross-condition preference optimization (Cross-CPO) to explicitly model the interactions between the two conditions. By ranking sentence-level and context-aware translations, CPL enables the model to learn a shared preference structure across conditions, allowing preference signals from one condition to influence learning in the other. Importantly, CPL operates entirely at the level of the training objective and requires no architectural changes, and can be applied to any shared-parameter translation system.

The main contributions in this paper can be summarized as follows:
\begin{itemize}[leftmargin=*]
\item We extend preference-based optimization to jointly support sentence-level and context-aware translation, providing a principled framework for multi-task preference learning.
\item We introduce CPL, which incorporates cross-condition preference optimization to explicitly model interactions between the two translation conditions and better exploit informative context.
\item We evaluate CPL on multiple document-level MT tasks and show that a single trained model performs robustly under both sentence-level and context-aware settings, achieving significant improvements in translation quality.
\end{itemize}

\section{Methodology}\label{sec:method}

\begin{figure*}[!t]
\centering
\includegraphics[width=0.95\textwidth, trim={0cm 0cm 0cm 0cm}]{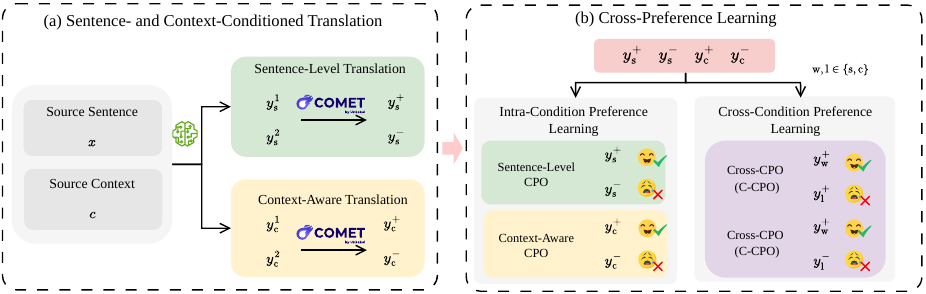}
\caption{Illustration of our approach.} 
\label{fig:method}
\end{figure*}

Figure~\ref{fig:method} provides an overview of Cross-Preference Learning (CPL). We first introduce the two translation settings in this work (Figure~\ref{fig:method}(a)) in Section~\ref{sec:two_tasks}, then describe our proposed CPL (Figure~\ref{fig:method}(b)) in Section~\ref{sec:cpl}.

\subsection{Sentence- and Context-Conditioned Translation}\label{sec:two_tasks}

We study a machine translation setting in which a single model operates under two source-side input conditions: sentence-level and context-aware translation. In the former, the model receives only a source sentence $x$, while in the latter it additionally receives source-side context $c$, such as neighboring sentences within a document.

The model, parameterized by $\theta$, defines two conditional distributions over target sentences: $p_\theta(y \mid x)$ for sentence-level translation and $p_\theta(y \mid x,c)$ for context-aware translation. For each distribution, we sample two candidates: $\{y_{\text{s}}^{1}, y_{\text{s}}^{2}\}$ from $p_\theta(y \mid x)$ and $\{y_{\text{c}}^{1}, y_{\text{c}}^{2}\}$ from $p_\theta(y \mid x, c)$.

For each candidate pair, we construct preference labels based on automatic evaluation. Taking sentence-level translation as an example, we score each candidate using the average of sentence-level COMET and document-level COMET (d-COMET)~\cite{vernikos-2022-embarrassingly} against the reference. The higher-scoring candidate is treated as the preferred translation $y_{\text{s}}^{+}$, while the lower-scoring one is treated as the dispreferred translation $y_{\text{s}}^{-}$. We apply the same procedure to the context-aware candidates to obtain the corresponding preference pair $(y_{\text{c}}^{+}, y_{\text{c}}^{-})$.

Since translations produced under the two conditions are directly comparable, this setup enables preference modeling both within and across conditions. Our goal is to train a single model that performs robustly under both settings.

\subsection{Cross-Preference Learning}\label{sec:cpl}
We first review preference-based optimization background (Section~\ref{sec:background}), then present intra- and cross-condition preference learning (Sections~\ref{sec:intra_cpo}--\ref{sec:cross_cpo}), and finally the overall CPL objective (Section~\ref{sec:cpl_objective}).

\subsubsection{Preference-Based Optimization Background}\label{sec:background}
Preference-based optimization trains models on pairwise comparisons of candidate translations rather than reference likelihood. For a preference pair $(y^+, y^-)$ and input $x$, the model is encouraged to assign higher probability to $y^+$. When context is available, we treat $x$ as $(x,c)$ implicitly.

We adopt contrastive preference optimization (CPO)~\cite{xu-2024-cpo}. The intra-condition CPO loss for a pair $(y^+,y^-)$ is:
\begin{equation}
\begin{aligned}
&\mathcal{L}_{\text{CPO}}(x,y^+,y^-)=\\[-2pt]
&\quad-\log \sigma\Bigl[\beta\bigl(
\log p_\theta(y^+\mid x)-\log p_\theta(y^-\mid x)
\bigr)\Bigr]\\[-2pt]
&\quad-\log p_\theta(y^+\mid x),
\end{aligned}
\label{equ:cpo}
\end{equation}
where $\sigma$ is the sigmoid function, $\beta$ controls preference separation.

\subsubsection{Intra-Condition Preference Learning} \label{sec:intra_cpo}

We extend contrastive preference optimization (CPO), originally formulated for a single input condition, to the sentence- and context-conditioned translation. This extension enables preference-based optimization to be applied consistently across the two related tasks within a shared-parameter model.

CPO is applied independently to the selected preference pairs for each condition: $(y_{\text{s}}^{+}, y_{\text{s}}^{-})$ for sentence-level translation under $p_\theta(y\mid x)$, and $(y_{\text{c}}^{+}, y_{\text{c}}^{-})$ for context-aware translation under $p_\theta(y \mid x, c)$. Let $\mathcal{P}_{\text{s}}$ and $\mathcal{P}_{\text{c}}$ denote the selected sets of sentence-level and context-aware preference pairs, respectively. Intra-condition preference learning minimizes the following objective:
{
\begin{align}
\mathcal{L}_{\text{intra}}(\theta) &=
\;\mathbb{E}_{(y^+,y^-)\sim \mathcal{P}_{\text{s}}}
\big[\mathcal{L}_{\text{CPO}}\left(x, y^+, y^-\right)\big] \notag \\
&\hspace{-2em}+ \mathbb{E}_{(y^+,y^-)\sim \mathcal{P}_{\text{c}}}
\big[\mathcal{L}_{\text{CPO}}\left(\left(x,c\right), y^+, y^-\right)\big],
\label{equ:intra_loss}
\end{align}
}

where $\mathcal{L}_{\text{CPO}}$ is the CPO loss defined in Eq.~\ref{equ:cpo}, computed over sentence-level and context-aware preference pairs, respectively.

The first expectation separates preferred and dispreferred candidates under the sentence-level distribution $p_\theta(y\mid x)$, while the second does the same under the context-aware distribution $p_\theta(y\mid x,c)$. Because the two terms share model parameters, preference signals are learned jointly across both input conditions.

The intra-condition loss $\mathcal{L}_{\text{intra}}(\theta)$ enables effective preference-based learning within each translation condition while sharing model parameters across both tasks. Appendix~\ref{apx:pair_selection} presents the details of preference pair selection for both {$\mathcal{P}_\text{s}$} and { $\mathcal{P}_\text{c}$}.

\subsubsection{Cross-Condition Preference Learning}\label{sec:cross_cpo}

Intra-condition preference learning optimizes sentence-level and context-aware translation independently, and therefore does not explicitly relate the two conditioned outputs. To exploit interactions across conditions, we introduce cross-condition preference learning, which directly compares sentence-level and context-aware translations for the same source sentence.

For each source sentence $x$ with context $c$, we identify the best translation $y_{\text{w}}^{+}$ from four candidates, i.e.,
$(y_{\text{s}}^{+}, y_{\text{s}}^{-})$ and $(y_{\text{c}}^{+}, y_{\text{c}}^{-})$.
The input condition that produces $y_{\text{w}}^{+}$ is defined as the winning condition ($\text{w}$), while the other serves as the losing condition ($\text{l}$), where $\text{w}, \text{l} \in \{\text{s}, \text{c}\}$. We denote the inputs corresponding to the winning and losing conditions as $x_{\text{w}}$ and $x_{\text{l}}$, respectively, where $x_{\text{w}}, x_{\text{l}} \in \{x_{\text{s}}, x_{\text{c}}\}$, with $x_{\text{s}}=x$ denoting the sentence-level input and $x_{\text{c}}=(x,c)$ denoting the context-aware input. 

Then we construct two cross-condition preference pairs: $(y_{\text{w}}^{+},y_{\text{l}}^{+})$ and $(y_{\text{w}}^{+},y_{\text{l}}^{-})$. These pairs encourage the model to consistently prefer the winning-condition output over both strong and weak outputs from the losing condition. For example, if $y_{\text{c}}^{+}$ is the best of the four candidates, then $\text{w}=\text{c}$, $\text{l}=\text{s}$, $y_{\text{w}}^{+}=y_{\text{c}}^{+}$, $x_{\text{w}}=(x,c)$, and $x_{\text{l}}=x$; the two cross-condition pairs are $(y_{\text{c}}^{+},y_{\text{s}}^{+})$ and $(y_{\text{c}}^{+},y_{\text{s}}^{-})$.

Let { $\mathcal{P}_{\text{cr}}$} denote the set of cross-condition preference pairs. The cross-condition preference loss is defined as:
{
\begin{align}
\label{equ:cross_loss}
\mathcal{L}_{\text{cross}}&(\theta) \\ =&\mathbb{E}_{(y_{\text{w}}^{+},y_{\text{l}}^{+})\sim \mathcal{P}_{\text{cr}}}\mathcal{L}_{\text{C-CPO}}\left(\{x_{\text{w}},x_{\text{l}}\}, y_{\text{w}}^{+}, y_{\text{l}}^{+}\right)\notag\\
 + &\mathbb{E}_{(y_{\text{w}}^{+},y_{\text{l}}^{-})\sim \mathcal{P}_{\text{cr}}}\mathcal{L}_{\text{C-CPO}}\left(\{x_{\text{w}},x_{\text{l}}\}, y_{\text{w}}^{+}, y_{\text{l}}^{-}\right) \notag.
\end{align}
}

The first expectation contrasts the globally best candidate with the stronger candidate from the other condition, yielding a fine-grained comparison. The second contrasts it with the weaker candidate, yielding a clearer preference relation with a larger quality gap. The cross-CPO (C-CPO) loss used in both expectations is defined as:
\begin{equation}
\begin{aligned}
&\mathcal{L}_{\text{C-CPO}}(\{x_{\text{w}},x_{\text{l}}\},y_{\text{w}},y_{\text{l}})=\\[-2pt]
&\quad-\log \sigma\Bigl[\beta\bigl(
\log p_\theta(y_{\text{w}}\mid x_{\text{w}})-\log p_\theta(y_{\text{l}}\mid x_{\text{l}})
\bigr)\Bigr]\\[-2pt]
&\quad-\log p_\theta(y_{\text{w}}\mid x_{\text{w}}).
\end{aligned}
\label{equ:new_cpo}
\end{equation}

As shown, cross-CPO adapts the standard CPO objective to allow comparisons between translations generated under different conditioning signals while preserving the contrastive learning objective. Since both conditions use the same model and output space, we argue that their likelihoods can be compared directly. This is further supported by our empirical analyses, which show closely aligned normalized log-probability scales and consistent gains from probability-based cross-condition reranking (Appendices~\ref{apx:calibration} and~\ref{apx:reranking}), supporting the practical validity of this approach. 

Overall, the cross-condition loss $\mathcal{L}_{\text{cross}}(\theta)$ encourages the model to assign higher probability to preferred translations across input conditions, explicitly modeling interaction between sentence-level and context-aware outputs. Appendix~\ref{apx:pair_selection} presents the details of preference pair selection for {\small $\mathcal{P}_\text{cr}$}.

\subsubsection{CPL objective}\label{sec:cpl_objective}
CPL unifies intra- and cross-condition preference signals into a single objective, jointly optimizing sentence-level and context-aware translation under different input conditions.

The CPL objective is defined as the sum of the corresponding preference-based losses defined in Eq.~\ref{equ:intra_loss} and Eq.~\ref{equ:cross_loss}:

\begin{equation} 
\mathcal{L}_{\text{CPL}}(\theta) = \mathcal{L}_{\text{intra}}(\theta) + \mathcal{L}_{\text{cross}}(\theta).
\label{equ:cpl_loss}
\end{equation}

By jointly optimizing intra- and cross-condition pairs, CPL enables a single model to perform robustly under both settings without architectural changes.

\section{Experimentation}\label{sec:experiment}

\begin{table*}[!t]
\centering
\small
\resizebox{\textwidth}{!}{
\begin{NiceTabular}[color-inside]{l|l|rcrcrcrcrcrc|rc}
\toprule
\Block[l]{2-1}{\textbf{\#}} & \Block[l]{2-1}{\textbf{Model}} & \Block[c]{1-2}{\textbf{En $\Rightarrow$ De}} &  & \Block[c]{1-2}{\textbf{En $\Rightarrow$ Es}} & & \Block[c]{1-2}{\textbf{En $\Rightarrow$ Fr}} & & \Block[c]{1-2}{\textbf{En $\Rightarrow$ It}} & & \Block[c]{1-2}{\textbf{En $\Rightarrow$ Nl}} && \Block[c]{1-2}{\textbf{En $\Rightarrow$ Ru}} & & \Block[c]{1-2}{\textbf{\textit{Average}}} &\\

\cmidrule(lr){3-4}\cmidrule(lr){5-6}\cmidrule(lr){7-8}\cmidrule(lr){9-10}\cmidrule(lr){11-12}\cmidrule(lr){13-14}\cmidrule(lr){15-16}

& & \textit{Sent} & \textit{CTX} & \textit{Sent} & \textit{CTX}  & \textit{Sent} & \textit{CTX}  & \textit{Sent} & \textit{CTX} & \textit{Sent} & \textit{CTX} & \textit{Sent} & \textit{CTX} & \textit{Sent} & \textit{CTX} \\

\rowcolor[gray]{0.85}
\Block[c]{1-16}{\texttt{Qwen3-4B}}\\

\cellcolor{sentlevel}1 & \cellcolor{sentlevel}Sent-Level & 82.51 & - & 84.91 & -  & 83.13 & -  & 86.94 & -  & 82.71 & -  & 83.12 & - & 83.89& -  \\
\cellcolor{sentlevel}2 & \cellcolor{sentlevel}Sent-Level$_\text{tuned}$  & 84.41 & - & 86.88  & -  & 84.62 & -  & 87.97 & -  & 84.88 & -  & 84.58 & - &  85.56 & -  \\
\cellcolor{sentlevel}3 & \cellcolor{sentlevel}Sent-Level$_\text{CPO}$ & 85.14 & - & 87.43 & -  & 85.18 & -  & 88.66 & -  & 85.88 & -  & 86.19 & - & 86.41 & -  \\
\cellcolor{ctxlevel}4 & \cellcolor{ctxlevel}Ctx-Aware  &- & 83.25 & -  & 86.98 & -  & 84.47 & -  & 87.81 & -  & 83.47 & - &  83.86 & - & 84.97 \\
\cellcolor{ctxlevel}5 & \cellcolor{ctxlevel}Ctx-Aware$_\text{tuned}$  & - & 84.31 & -  & 86.85 & -  & 84.64 & -  & 88.01 & -  & 84.85 & - &  84.86 & - & 85.59 \\
\cellcolor{ctxlevel}6 & \cellcolor{ctxlevel}Ctx-Aware$_\text{CPO}$  & - & 85.38 & -  & 87.56 & -  & 85.32 & -  & 88.78 & -  & 85.91 & - & 86.20 & - & 86.52 \\
\cellcolor{sentctxlevel}7 & \cellcolor{sentctxlevel}Sent+Ctx$_{\text{tuned}}$ 
& 84.66 & 84.79 & 87.07 & 87.02 & 84.60 & 84.61 & 88.18 & 88.27 & 85.16 & 85.08 & 85.04 & 84.95 & 85.79 & 85.79 \\
\cellcolor{sentctxlevel}8 & \cellcolor{sentctxlevel}Sent+Ctx$_{\text{tuned+}}$ & 84.58 & 84.60 & 87.11 & 87.12 & 84.74 & 84.78 & 88.39 & 88.39 & 85.31 & 85.32 & 84.88 & 84.83 & 85.84 & 85.84 \\
\cellcolor{our}9 & \cellcolor{our}CPL (Ours)  & \cellcolor{darkred!80}85.92 & \cellcolor{darkred!80}85.91 & \cellcolor{darkred!80}87.75 & \cellcolor{darkred!80}87.86 & \cellcolor{darkred!80}85.84 & \cellcolor{darkred!80}85.82 & \cellcolor{darkred!80}89.08 & \cellcolor{darkred!80}89.06 & \cellcolor{darkred!80}86.52 & \cellcolor{darkred!80}86.57 & \cellcolor{darkred!80}86.50 & \cellcolor{darkred!80}86.56 & \cellcolor{darkred!80}86.94 & \cellcolor{darkred!80}86.96 \\

\cellcolor{our}10 & \cellcolor{our}~~ - IntraLoss & 85.53 & 85.53 & \cellcolor{lightred!80}87.63 & 87.63 & \cellcolor{lightred!80}85.56 & 85.52 & \cellcolor{lightred!80}88.92 & \cellcolor{lightred!80}88.95 & 86.20 & 86.21 & 86.09 & 85.97 & 86.66 & 86.64 \\
\cellcolor{our}11 & \cellcolor{our}~~ - CrossLoss & \cellcolor{lightred!80}85.58 & \cellcolor{lightred!80}85.66 & 87.59 & \cellcolor{lightred!80}87.73 & 85.54 & \cellcolor{lightred!80}85.66 & 88.81 & 88.91 & \cellcolor{lightred!80}86.46 & \cellcolor{lightred!80}86.46 & \cellcolor{lightred!80}86.41 & \cellcolor{lightred!80}86.49 & \cellcolor{lightred!80}86.73 & \cellcolor{lightred!80}86.82 \\

\rowcolor[gray]{0.85}
\Block[c]{1-16}{\texttt{Qwen3-8B}}\\
\cellcolor{sentlevel}1 & \cellcolor{sentlevel}Sent-Level & 85.13 & - &87.59 & - &85.25 & - &88.70 & - &85.93 & - &85.76 & - & 86.39 & - \\
\cellcolor{sentlevel}2 & \cellcolor{sentlevel}Sent-Level$_\text{tuned}$ &86.12 & - &87.73 & - & 85.82& - &89.12 & - &87.07 & - &86.34 & - &87.03 & - \\
\cellcolor{sentlevel}3 & \cellcolor{sentlevel}Sent-Level$_\text{CPO}$ & 86.26 & - & 88.20 & -  & 86.14 & -  & 89.44 & -  & 87.67 & -  & 87.33 & - & 87.51 & -  \\
\cellcolor{ctxlevel}4 & \cellcolor{ctxlevel}Ctx-Aware & - & 85.05 & - &87.72 & - & 85.35& - &88.75 & - &86.09 & - &85.59 & - & 86.43\\
\cellcolor{ctxlevel}5 & \cellcolor{ctxlevel}Ctx-Aware$_\text{tuned}$ & - &86.13 & - &87.78 & - &85.80 & - &88.99 & - &87.11 & - &86.39 & - & 87.03 \\
\cellcolor{ctxlevel}6 & \cellcolor{ctxlevel}Ctx-Aware$_\text{CPO}$  & - & 86.54 & -  & 88.29 & -  & 86.26 & -  & 89.47 & -  & 87.60 & - & 87.32 & - & 87.58 \\
\cellcolor{sentctxlevel}7 & \cellcolor{sentctxlevel}Sent+Ctx$_{\text{tuned}}$ & 86.14 & 86.24 &87.82 & 87.86 &85.71 & 85.78 &89.16 & 89.22 &87.30 & 87.33 &86.51 & 86.47 & 87.11 & 87.15 \\
\cellcolor{sentctxlevel}8 & \cellcolor{sentctxlevel}Sent+Ctx$_{\text{tuned+}}$ & 86.24 & 86.34 & 87.81 & 87.94 & 85.80 & 85.89 & 89.25 & 89.23 & 87.31 & 87.47 & 86.46 & 86.29 & 87.15 & 87.19
\\
\cellcolor{our}9 & \cellcolor{our}CPL (Ours) & \cellcolor{darkred!80}87.19 & \cellcolor{darkred!80}87.23  & \cellcolor{darkred!80}88.47 & \cellcolor{darkred!80}88.51  & \cellcolor{darkred!80}86.56 & \cellcolor{darkred!80}86.63  & \cellcolor{darkred!80}89.81 & \cellcolor{darkred!80}89.99  & \cellcolor{darkred!80}88.22 & \cellcolor{darkred!80}88.28  & \cellcolor{darkred!80}87.71 & \cellcolor{darkred!80}87.75 & \cellcolor{darkred!80}87.99  & \cellcolor{darkred!80}88.06  \\
\cellcolor{our}10 & \cellcolor{our}~~ - IntraLoss & 86.85 & 86.85 & \cellcolor{lightred!80}88.34 & 88.35  & 86.38 &  86.34 & 89.68 & 89.67  &87.86  & 87.83  & \cellcolor{lightred!80}87.41 &87.26  & 87.75  & 87.72  \\
\cellcolor{our}11 & \cellcolor{our}~~ - CrossLoss & \cellcolor{lightred!80}86.88 & \cellcolor{lightred!80}87.00  & 88.33  & \cellcolor{lightred!80}88.39  & \cellcolor{lightred!80}86.39 & \cellcolor{lightred!80}86.54  & \cellcolor{lightred!80}89.74 & \cellcolor{lightred!80}89.74  & \cellcolor{lightred!80}87.97 & \cellcolor{lightred!80}87.99  & 87.35 & \cellcolor{lightred!80}87.43  & \cellcolor{lightred!80}87.78  & \cellcolor{lightred!80}87.85  \\
\rowcolor[gray]{0.85}
\Block[c]{1-16}{\texttt{Llama-3-8B-Instruct}}\\
\cellcolor{sentlevel}1 & \cellcolor{sentlevel}Sent-Level & 82.72 & - & 84.40 & -  & 83.07 & -  & 86.11 & -  & 85.20 & -  &	82.15  & - & 83.94 & -  \\
\cellcolor{sentlevel}2 & \cellcolor{sentlevel}Sent-Level$_\text{tuned}$  & 86.00 & - & 87.59 & -  & 85.44 & -  & 88.99 & -  & 88.22 & -  & 85.82  & - & 87.01 & -  \\
\cellcolor{sentlevel}3 & \cellcolor{sentlevel}Sent-Level$_\text{CPO}$ & 86.49 & - & 88.10 & -  & 85.92 & -  & 89.55 & -  & 88.63 & -  & 86.71 & - & 87.57 & -  \\
\cellcolor{ctxlevel}4 & \cellcolor{ctxlevel}Ctx-Aware & - & 51.33  & -  & 56.62  & -  & 50.41 & -  & 58.05 & -  & 53.51 & - & 53.05  & - & 53.83 \\
\cellcolor{ctxlevel}5 & \cellcolor{ctxlevel}Ctx-Aware$_\text{tuned}$ & - & 86.10 & -  & 87.67 & -  & 85.54 & -  & 89.10 & -  &88.28  & - & 85.89  & - &87.10 \\
\cellcolor{ctxlevel}6 & \cellcolor{ctxlevel}Ctx-Aware$_\text{CPO}$  & - & 86.72 & -  & 88.15 & -  & 86.19 & -  & 89.56 & -  & 88.69 & - & 86.77 & - & 87.68 \\
\cellcolor{sentctxlevel}7 & \cellcolor{sentctxlevel}Sent+Ctx$_{\text{tuned}}$ & 86.42 & 86.56  & 87.84 & 87.85 & 85.76 & 85.85	& 89.28 &89.31 	&88.61& 88.62	&86.40& 86.17 & 87.39 & 87.39 \\
\cellcolor{sentctxlevel}8 & \cellcolor{sentctxlevel}Sent+Ctx$_{\text{tuned+}}$ & 86.43 & 86.51 & 87.92 & 87.93 & 85.95 & 85.97 & 89.42 & 89.37 & 88.74 & 88.75 & 86.26 & 86.23 & 87.45 & 87.46 \\
\cellcolor{our}9 & \cellcolor{our}CPL (Ours)& \cellcolor{darkred!80}87.09 & \cellcolor{darkred!80}87.19 &\cellcolor{darkred!80} 88.43 &\cellcolor{darkred!80} 88.45 &\cellcolor{darkred!80} 86.60 &\cellcolor{darkred!80} 86.65 & \cellcolor{darkred!80}89.82 &\cellcolor{darkred!80} 89.85 &\cellcolor{lightred!80} 88.93 &\cellcolor{darkred!80} 89.00 & \cellcolor{darkred!80}87.42 &\cellcolor{darkred!80} 87.43 &\cellcolor{darkred!80} 88.05 &\cellcolor{darkred!80} 88.10 \\
\cellcolor{our}10 & \cellcolor{our}~~ - IntraLoss & \cellcolor{lightred!80}87.00 & 87.03 & \cellcolor{lightred!80}88.29 & \cellcolor{lightred!80}88.32 & \cellcolor{lightred!80}86.44 & \cellcolor{lightred!80}86.47 & \cellcolor{lightred!80}89.69 & \cellcolor{lightred!80} 89.77 & 88.79 & 88.79 & 87.16 & 86.94 & 87.90 & 87.89 \\
\cellcolor{our}11 & \cellcolor{our}~~ - CrossLoss & 86.98 &\cellcolor{lightred!80} 87.11 & 88.17 & 88.16 & 86.41 & 86.42 &\cellcolor{lightred!80} 89.69 & 89.71 &\cellcolor{darkred!80} 88.94 & \cellcolor{lightred!80}88.95 & \cellcolor{lightred!80}87.35 & \cellcolor{lightred!80}87.39 & \cellcolor{lightred!80}87.92 &\cellcolor{lightred!80} 87.96 \\
\bottomrule
\end{NiceTabular}
}
\caption{Performance comparison on COMET. Best and second-best scores are highlighted in \colorbox{darkred!80}{dark red} and \colorbox{lightred!80}{light red}. \textit{Sent} and \textit{CTX} denote sentence-level input ($x$) and context-aware input ($x$ and context $c$) during inference.}
\label{tab:comet}
\end{table*}

\subsection{Experimental Settings}\label{sec:setting}
%
\noindent\textbf{Datasets.} Following recent work, we use the document-level parallel data from News Commentary v18.1 released in WMT25,\footnote{\url{https://www2.statmt.org/wmt25/translation-task.html}} which contains document boundary annotations. Our main experiments cover six English-to-X directions (En$\Rightarrow$De/Es/Fr/It/Nl/Ru) across all three backbones. We also evaluate Qwen3-8B on five X$\Rightarrow$En directions (Es/Fr/Nl/Ru/Zh) from the same corpus, and four IWSLT2017 directions (En$\Rightarrow$De/Zh/Fr and Zh$\Rightarrow$En), whose spoken-language documents exhibit stronger discourse dependencies.\footnote{\url{https://huggingface.co/datasets/IWSLT/iwslt2017/tree/main/data}} More dataset details are provided in Appendix~\ref{apx:dataset}.

Similar to \citet{xu-2024-cpo}, we constructed a small-scale dataset for the cold-start fine-tuning stage. For each direction, we extract a contiguous block of 20,000 sentences sequentially, ensuring the availability of context for context-aware training. Specifically, for the context-aware samples, we prepend the preceding source text, truncated to the last 256 tokens \cite{lyu-2024-dempt}, as the input context. 

Following \citet{xu-2024-cpo} and \citet{wang-2026-m2po}, we construct the preference pairs from high-quality validation sets. In our setup, these validation sets do not overlap with any test set. For each source sentence in the validation sets, the jointly fine-tuned model generates two candidates per input condition, using greedy decoding and sampling at temperature 0.7, respectively. Appendix~\ref{apx:dataset} shows the statistics of datasets used for CPL learning and the preference pairs used.  

\noindent\textbf{Models and Settings.} We select Qwen3-4B\footnote{\url{https://huggingface.co/Qwen/Qwen3-4B}}~\cite{yang-2025-qwen3}, Qwen3-8B\footnote{\url{https://huggingface.co/Qwen/Qwen3-8B}} and Meta Llama-3-8B-Instruct\footnote{\url{https://huggingface.co/meta-llama/Meta-Llama-3-8B-Instruct}}~\cite{meta-2024-llama3} as the foundation open-source LLMs. For detailed fine-tuning and hyper-parameter settings, please refer to Appendix~\ref{apx:settings}.

\noindent\textbf{Baselines.}
We compare our approach against the following baselines:

\begin{itemize}[leftmargin=*]
\item \textbf{Sent-Level.} The base LLM translates each sentence independently. \textbf{Sent-Level$_{\text{tuned}}$} is fine-tuned on sentence-level data. We additionally apply CPO to this model using sentence-level preference pairs, yielding \textbf{Sent-Level$_{\text{CPO}}$}.

\item \textbf{Ctx-Aware.} The base LLM translates using the current sentence and its preceding context. \textbf{Ctx-Aware${_\text{tuned}}$} is fine-tuned on context-aware data. Similarly, \textbf{Ctx-Aware${_\text{CPO}}$} applies CPO with context-aware preference pairs.
\item \textbf{Sent+Ctx${_\text{tuned}}$.} A naive multi-condition baseline jointly fine-tuned on both sentence-level and context-aware data using standard likelihood training. \textbf{Sent+Ctx${_\text{tuned+}}$} further incorporates the dataset used for extracting CPL preference pairs during fine-tuning.
\end{itemize}

Note that, for fair comparison, Sent+Ctx$_{\text{tuned+}}$, Sent-Level$_{\text{CPO}}$, and Ctx-Aware$_{\text{CPO}}$ all use the same additional data (i.e., the validation sets) as CPL. Both Sent+Ctx$_{\text{tuned}}$ and CPL support two input conditions at inference time: (i) sentence-level translation with input $x$, and (ii) context-aware translation with input $(x, c)$. Detailed prompt templates are provided in Appendix~\ref{apx:prompts}, and the inference cost is reported in Appendix~\ref{apx:inference_efficiency}.

\noindent\textbf{Metrics.}
We primarily report \textit{sentence-level COMET} scores~\cite{rei-2020-comet}. Specifically, we use the reference-based metric \texttt{wmt22-comet-da}~\cite{rei-2022-comet}. In addition, we report supplementary evaluation metrics, including sentence-level BLEU (s-BLEU)~\cite{post-2018-call} and document-level COMET (d-COMET), in Appendix~\ref{apx:more_results}. Since our model and most baselines support both input conditions, we evaluate and report performance for both sentence-level (\textit{Sent}) and context-aware (\textit{CTX}) translation.

\subsection{Main Results}
Table~\ref{tab:comet} reports the results on En-to-X translation directions from News Commentary. Results for other translation directions are provided in Appendix~\ref{apx:en_to_x} and Appendix~\ref{apx:iwslt2017}. Across all models and training settings, sentence-level and context-aware translations (\#2 vs. \#5) achieve very similar COMET scores, suggesting that document-level context does not consistently improve average translation quality. Similar observations have been reported in prior work, where contextual benefits are often sparse and localized~\cite{dong-2025-two,lyu-2024-dempt}. This motivates CPL: we aim to learn preference signals that better exploit informative context. Based on Table~\ref{tab:comet}, we further observe:
\begin{itemize}[leftmargin=*]
\item \textbf{Effect of fine-tuning and joint training.}
Among the baselines, fine-tuned LLMs consistently outperform their untuned counterparts (e.g., \#2 vs. \#1 and \#5 vs. \#4). Joint fine-tuning on both sentence- and context-conditioned data further improves both tasks (\#7 vs. \#2 and \#7 vs. \#5). For example, with Llama-3-8B-Instruct, joint training yields average COMET gains of 0.38 and 0.29 over individually fine-tuned models, indicating that we establish a strong and competitive baseline.
\item \textbf{Impact of incorporating additional training data.}
Adding the CPL data during joint fine-tuning (Sent+Ctx$_{\text{tuned+}}$, \#8) yields only marginal gains, suggesting that performance gains are not simply due to more data, but require more effective learning signals. In contrast, applying CPO with the same data, both Sent-Level$_\text{CPO}$ (\#3) and Ctx-Aware$_\text{CPO}$ (\#6) achieve larger improvement for both sentence-level and context-aware translation, respectively. This highlights the value of preference-based learning.
\item \textbf{Effectiveness of CPL.}
Compared with all baselines, CPL (\#9) achieves substantial improvements. Notably, when compared to Sent+Ctx$_{\text{tuned+}}$ (\#8), which leverages the same training data, CPL yields consistent gains for both sentence-level and context-aware translation. For instance, CPL improves average COMET by 1.10/1.12, 0.84/0.87, and 0.60/0.64 for sentence-level/context-aware translation using Qwen3-4B, Qwen3-8B, and Llama-3-8B-Instruct, respectively. This highlights the effectiveness of Cross-Preference Learning beyond conventional fine-tuning. Moreover, CPL (\#9) outperforms the two other strong baselines Sent-Level$_\text{CPO}$ (\#3) and Ctx-Aware$_\text{CPO}$ (\#6) which are augmented with preference learning on the same training data. 
\item \textbf{Ablation studies.}
Ablation results (\#9 vs. \#10 vs. \#11) show that both intra-condition loss and cross-condition loss contribute to CPL's performance. Removing either component degrades performance, with intra-condition loss having a slightly larger impact, indicating its complementary role in stabilizing preference learning.
\item \textbf{Consistency across model scales.} 
Finally, the observed performance trends are consistent across all three LLM backbones, suggesting that CPL is robust and model-agnostic, and can be effectively applied to different pretrained models.
\end{itemize}

\section{Discussion} \label{sec:discussion}

In this section, we use Qwen3-8B as the foundation model to discuss our approach. Unless otherwise specified, we report the average sentence-level COMET scores across all six translation tasks. Additional analyses, including the effects of context length, human evaluation, and case studies, are provided in Appendix~\ref{apx:context_length}--\ref{apx:case_study}.

\subsection{Can CPL Better Exploit Informative Context?}\label{sec:cpl_learn}
We examine whether CPL can exploit useful context while remaining robust to noisy context. Using the En$\Rightarrow$De test set from News Commentary, we ask GPT-5 (GPT-5-2025-08-07)~\cite{openai-2025-gpt5} to label whether the source-side context is helpful for translating each sentence. We retain 696 out of 5,967 sentences labeled as \textit{High}, indicating that context is highly necessary for accurate translation.\footnote{See Appendix~\ref{apx:identifying_context} for labeling details.} We further evaluate on the En$\Rightarrow$De test set from IWSLT, where discourse dependencies are generally stronger, selecting 2,183 out of 8,079 sentences using the same procedure.

Table~\ref{tab:context_usefulness} shows that when the original context is used (Ctx-gold), context-aware CPL consistently outperforms sentence-level CPL by 0.21 and 0.64 d-COMET on the two test sets, suggesting that the model effectively leverages informative context. When the context is replaced with randomly sampled sentences from other documents (Ctx-random), context-aware CPL performs nearly identically to sentence-level translation. This trend is consistent across both datasets and is more pronounced on IWSLT, where contextual dependencies are stronger. Overall, the results suggest that CPL can benefit from useful context while remaining robust to noisy context.

\begin{table}[t]
\centering
\small
\begin{tabular}{lccc}
\toprule
\textbf{Dataset} & \textbf{Sent} & \textbf{Ctx-gold} & \textbf{Ctx-random} \\
\midrule
NewsCommentary & 82.50 & 82.71 & 82.52 \\
IWSLT & 78.06 & 78.70 & 78.02 \\
\bottomrule
\end{tabular}
\caption{Performance (document-level COMET) on sentences with useful context. \textit{Ctx-random} replaces the original context with randomly sampled one.}
\label{tab:context_usefulness}
\end{table}

\subsection{Analysis with Additional Quality Metrics}\label{sec:additional_metrics}
Beyond BLEU and COMET, we evaluate CPL using complementary metrics targeting discourse coherence, faithfulness, and fluency, which are particularly important for context-aware translation.

Following \citet{li-2023-contrastive} and \citet{dong-2025-two}, we measure discourse coherence (Coh.) using cosine similarity between neighboring sentence embeddings from SimCSE~\cite{gao-2021-simcse}. Higher scores indicate better consistency across sentence boundaries. For faithfulness, we report ALTI+~\cite{ferrando-2022-towards}, which uses a pretrained NLLB model to detect under-translation and hallucination.\footnote{\url{https://facebookresearch.github.io/stopes/docs/eval/alti}} Higher ALTI+ scores correspond to fewer faithfulness issues. Finally, we follow \citet{sun-2025-fine} and evaluate translation fluency using GPT-5 (GPT-5-2025-08-07)~\cite{openai-2025-gpt5} as an external judge.\footnote{Due to the high API cost of fluency evaluation (about \$50 per direction), we report averaged results on En$\Rightarrow$De and En$\Rightarrow$Ru only.} Among them, Coh. and Fluency are document-level metrics requiring document-level context.

Table~\ref{tab:additional_metrics} shows that Sent+Ctx$_{\text{tuned}}$ improves coherence, faithfulness, and fluency over the base model, indicating more effective use of document context. CPL further achieves the best performance across all metrics, with higher discourse coherence, fewer faithfulness issues as measured by ALTI+, and improved fluency. These consistent gains suggest that CPL enhances translation quality beyond standard COMET evaluation.

\begin{table}[!t]
\centering
\small
\setlength{\tabcolsep}{4pt}
\begin{tabular}{l|ccc}
\toprule
\textbf{Model} & \textbf{Coh.} $\uparrow$ & \textbf{ALTI+} $\uparrow$ & \textbf{Fluency} $\uparrow$ \\
\midrule
Qwen3-8B & 73.28 & 54.49 & 3.81 \\
Sent+Ctx$_{\text{tuned}}$ & \cellcolor{lightred!80}74.78 & \cellcolor{lightred!80}54.67 & \cellcolor{lightred!80}3.90 \\
CPL (ours) & \cellcolor{darkred!80}75.24 & \cellcolor{darkred!80}54.90 & \cellcolor{darkred!80}4.04 \\
\bottomrule
\end{tabular}
\caption{Additional quality analysis using coherence (Coh.), ALTI+, and GPT-based fluency.}
\label{tab:additional_metrics}
\end{table}

To further evaluate discourse-specific phenomena, we additionally report BlonD-D~\cite{jiang-2022-blonde}, the discourse component of BlonDe, which aggregates category-level scores over entities, tense, pronouns, and discourse markers (DM). Since the BlonDe toolkit\footnote{\url{https://github.com/EleanorJiang/BlonDe}} currently supports only X$\Rightarrow$En translation directions, we report averages for Qwen3-8B-based systems with context-aware input on the News Commentary X$\Rightarrow$En tasks.

\begin{table}[!t]
\centering
\small
\setlength{\tabcolsep}{3pt}
\resizebox{\columnwidth}{!}{%
\begin{tabular}{l|ccccc}
\toprule
\textbf{Model} & \textbf{Entity} & \textbf{Tense} & \textbf{Pronoun} & \textbf{DM} & \textbf{BlonD-D} \\
\midrule
Qwen3-8B & 61.71 & 65.44 & 74.41 & 58.01 & 64.66 \\
Sent+Ctx$_{\text{tuned}}$ & \cellcolor{lightred!80}77.46 & \cellcolor{lightred!80}77.10 & \cellcolor{lightred!80}82.24 & \cellcolor{lightred!80}76.76 & \cellcolor{lightred!80}78.35 \\
CPL (ours) & \cellcolor{darkred!80}78.22 & \cellcolor{darkred!80}77.24 & \cellcolor{darkred!80}82.38 & \cellcolor{darkred!80}76.99 & \cellcolor{darkred!80}78.67 \\
\bottomrule
\end{tabular}
}
\caption{Average BlonD-D scores on News Commentary X$\Rightarrow$En tasks with context-aware translation.}
\label{tab:blonde}
\end{table}

Table~\ref{tab:blonde} shows that CPL outperforms Sent+Ctx$_{\text{tuned}}$ across every evaluated discourse phenomenon, suggesting improved use of contextual information beyond COMET-based evaluation.

\subsection{Evaluation with GPT-as-Judge}\label{sec:gpt_as_judge}

To obtain a complementary quality assessment beyond automatic metrics, we conduct an evaluation using GPT-as-Judge. We randomly sample 1,200 sentences from the test set. For each instance, we provide GPT-5 with the source sentence, its source-side context, and the reference translation. We then present four candidate translations in random order: the sentence-level and context-aware outputs from both Sent+Ctx$_{\text{tuned}}$ and CPL (i.e., systems \#7 and \#9 in Table~\ref{tab:comet}). The order of candidates is randomized to mitigate positional bias.

GPT-5 is asked to score each translation along four dimensions: \textit{contextual coherence}, \textit{fluency}, \textit{faithfulness}, and \textit{terminology expression}, each on a 0-25 scale, with the overall score computed as their sum. The evaluation prompt is provided in Figure~\ref{fig:gpt_evaluation} in Appendix~\ref{apx:GPT-as-Judge}. We report the average overall scores across all evaluated sentences.

As shown in Table~\ref{tab:evaluation_gpt}, CPL consistently outperforms Sent+Ctx$_{\text{tuned}}$ under both sentence-level and context-aware settings, with gains of +1.99 and +1.58, respectively. These improvements indicate that CPL enhances translation quality across multiple dimensions. Importantly, the gains are observed under both input conditions, suggesting that CPL provides overall quality improvements rather than benefiting only from traditional automatic metrics. Furthermore, the context-aware variant of CPL slightly outperforms its sentence-level counterpart, achieving the highest overall score. At the instance level, the context-aware CPL produces the best translation for 30.0\% of the evaluated sentences, compared to 26.8\% for sentence-level CPL.

Appendix~\ref{apx:human_eval} (Study 2) presents a human evaluation on 200 randomly sampled test cases, validating the reliability of the GPT-as-Judge evaluation framework.
\begin{table}[!t]
\centering
\small
\begin{tabular}{l|cc}
\toprule
\textbf{Model} & \textbf{Sent} & \textbf{CTX} \\
\midrule
Sent+Ctx$_{\text{tuned}}$ & 85.74 & 86.31 \\
CPL & 87.73 & 87.89 \\
\bottomrule
\end{tabular}
\caption{Evaluation results of GPT-as-Judge.}
\label{tab:evaluation_gpt}
\end{table}

\subsection{Effect of Preference Selection Metrics}
Cross-Preference Learning relies on ranking candidate translations to construct informative intra-condition and cross-condition preference pairs. By default, we use the average of document-level COMET (d-COMET) and sentence-level COMET (s-COMET) as the selection metric. To assess the sensitivity of CPL to this design choice, we compare several alternatives, including d-COMET alone, s-COMET alone, and BLEU~\cite{post-2018-call}.

Table~\ref{tab:metric_ablation} shows the performance. Among the COMET-based variants, performance differences are negligible, indicating that CPL is robust to different granularities of COMET scoring. In contrast, using BLEU as the selection metric leads to a noticeable performance drop. This is expected, as BLEU relies on surface-level n-gram overlap and is less aligned with semantic adequacy and discourse-level quality, which are crucial for selecting informative preference pairs.

\begin{table}[!t]
\centering
\small
\setlength{\tabcolsep}{4pt}
\begin{tabular}{l|ll}
\toprule
\textbf{Metric Used} & \textbf{Sent} & \textbf{CTX} \\
\midrule
BLEU          & 86.96 & 87.16 \\
d-COMET       & 87.97 & 87.98 \\
s-COMET       & 87.97 & 88.03 \\
(d- + s-COMET)/2 & 87.99 & 88.06 \\
\bottomrule
\end{tabular}
\caption{CPL performance (in COMET) using different metrics for selecting intra- and cross-condition preference pairs.}
\label{tab:metric_ablation}
\end{table} 

Because COMET-style metrics are used for both preference construction and primary evaluation, some metric alignment is unavoidable. We mitigate this concern with independent GPT-as-Judge (Section~\ref{sec:gpt_as_judge}) and human evaluations (Study 2 in Appendix~\ref{apx:human_eval}), human verification of preference pairs (Study 1 in Appendix~\ref{apx:human_eval}), and complementary discourse-level metrics (BlonD-D and ALTI+) (Section~\ref{sec:additional_metrics}), which consistently support the effectiveness of CPL.

\subsection{Effect of Cross-Condition Decomposition}

In cross-condition preference learning, we include two types of preference constraints: $(y_{\text{w}}^{+},y_{\text{l}}^{+})$ and $(y_{\text{w}}^{+},y_{\text{l}}^{-})$. To analyze their respective contributions, we conduct an ablation study in which we remove one of the two preference pairs from the cross-condition loss. 

As shown in Table~\ref{tab:cross_pair_ablation}, removing either constraint causes a small but consistent performance drop for both sentence-level and context-aware translation, indicating that both are beneficial. The degradation is slightly larger when removing $(y_{\text{w}}^{+}, y_{\text{l}}^{-})$, suggesting that contrasting against a weaker losing candidate provides a marginally stronger signal. These results suggest that the two cross-condition constraints play complementary roles and that CPL is not overly sensitive to any single preference relation. This redundancy contributes to the robustness of CPL, allowing CPL to maintain stable performance even when individual cross-condition signals are partially removed.

\begin{table}[!t]
\centering
\small
\begin{tabular}{l|cc}
\toprule
\textbf{Model} & \textbf{Sent} & \textbf{CTX} \\
\midrule
CPL &87.99 & 88.06 \\
CPL w/o $(y_{\text{w}}^{+},y_{\text{l}}^{+})$ & 87.98 &  88.01\\
CPL w/o $(y_{\text{w}}^{+},y_{\text{l}}^{-})$ & 87.88 & 87.91 \\
\bottomrule
\end{tabular}
\caption{Effect of removing individual cross-condition preference constraints.}
\label{tab:cross_pair_ablation}
\end{table}

\section{Related Work}\label{sec:related_work}

\noindent\textbf{LLM-based Document-level Machine Translation.}
With the emergence of LLMs, DocMT has been revisited by leveraging LLMs' strong contextual modeling and long-context capabilities. Early studies primarily evaluated prompt-based LLM performance on DocMT, analyzing how models such as ChatGPT handle extended source-side context \citep{wang-2023-document-level, karpinska-iyyer-2023-large}. Subsequent work proposes more structured prompting and inference-time strategies, including retrieval-augmented generation \citep{cui-2024-efficiently}, agent-based frameworks with memory mechanisms \citep{wang-2025-delta, guo-2025-doc}, observe-and-act agent pipelines that adaptively select context segment by segment as translation proceeds \citep{wang-2026-loong}, and iterative translation schemes that preserve discourse information across segments \citep{hu-2025-source}.

Another line of work adapts LLMs to DocMT via supervised fine-tuning (SFT), including instruction mixing \citep{li-2024-enhancing}, multi-stage fine-tuning \citep{wu-2024-adapting}, document-level monolingual post-editing of NMT outputs~\citep{li-2025-enhancing}, document-level evaluation of sentence-level SFT \citep{stap-2024-fine}, and decoding/prompt-enhanced strategies \citep{lyu-2024-dempt}. \citet{dong-2025-two} further refine translations by combining sentence- and document-level outputs. \citet{ramos-2025-docblocks} fine-tune LLMs with mixed context/no-context instructions. In a complementary analysis, \citet{mohammed-2025-context} show that document-level gains are inconsistent on discourse phenomena, suggesting context should be exploited selectively.

Unlike these studies, our work explicitly models interactions between sentence-level and context-aware translation outputs to learn preference signals that better exploit informative context.

\noindent\textbf{Preference Learning for Machine Translation.}
Preference learning has long been explored in MT as an alternative to likelihood-based training~\cite{hopkins-may-2011-tuning,cherry-foster-2012-batch}, using ranking-based or pairwise objectives to better align outputs with human judgments. Recent advances in LLM alignment, including RLHF \citep{christiano-2017-deep}, Direct Preference Optimization (DPO; \citealp{rafailov-2023-dpo}), and Contrastive Preference Optimization (CPO; \citealp{xu-2024-cpo}), have renewed interest in this paradigm. Several studies adapt these techniques to MT by constructing preference pairs from human annotations or automatic metrics, including quality-estimation models used as a proxy for human preference~\citep{uhlig-2025-dqo}, showing improvements in translation adequacy and fluency~\citep{agrawal-2024-modeling,wu-2024-word,yang-2024-direct,cui-2025-crpo,sun-2025-enhancing}. In contrast to prior work, which applies preference learning within a single translation setting, our approach extends preference learning to a dual-condition scenario by explicitly modeling interactions between sentence-level and context-aware translation.

\section{Conclusion}
In this paper, we study sentence-level and context-aware machine translation from a unified perspective and show that, while their overall performance is often similar, their outputs exhibit substantial and systematic differences. Motivated by this observation, we propose Cross-Preference Learning (CPL), a preference-based optimization framework that jointly models intra-condition and cross-condition translation preferences, enabling a single model to make better use of informative document context. Experiments across multiple document-level translation tasks demonstrate that CPL consistently improves both sentence-level and context-aware translation quality, enhances cross-condition consistency, and remains robust to context noise and metric choices. Our analysis further shows that CPL provides a simple yet effective way to align translation behavior across input conditions without architectural changes, offering a promising direction for robust multi-condition machine translation.

\section*{Limitations}
While Cross-Preference Learning (CPL) shows consistent improvements for both sentence-level and context-aware translation, several limitations remain. First, since CPL constructs preference pairs using automatic metrics and COMET is also our primary evaluation metric, some degree of metric-induced bias is unavoidable. Independent GPT-as-Judge and human evaluations, together with BlonD-D and ALTI+, mitigate but do not eliminate this concern. Second, CPL does not include an explicit mechanism that decides whether to use context; instead, a single model is implicitly encouraged to handle both input conditions within a unified preference-learning framework. Since the model remains a black box, we cannot determine which parts of the context are actually utilized, and we leave the interpretability of context usage to future work. Third, the ability of CPL to handle extremely long documents or complex discourse phenomena has not been fully explored. 

\section*{Acknowledgements}
We use AI assistants solely to polish writing and check syntax errors. We would like to thank the anonymous reviewers for their constructive feedback. 

\bibliography{main}

\begin{thebibliography}{46}
\providecommand{\natexlab}[1]{#1}

\bibitem[{Agrawal et~al.(2024)Agrawal, De~Souza, Rei, Farinhas, Faria, Fernandes, Guerreiro, and Martins}]{agrawal-2024-modeling}
Sweta Agrawal, Jos{\'e} G.~C. De~Souza, Ricardo Rei, Ant{\'o}nio Farinhas, Gon{\c{c}}alo Faria, Patrick Fernandes, Nuno~M Guerreiro, and Andre Martins. 2024.
\newblock \href {https://doi.org/10.18653/v1/2024.emnlp-main.803} {Modeling user preferences with automatic metrics: Creating a high-quality preference dataset for machine translation}.
\newblock In \emph{Proceedings of EMNLP}, pages 14503--14519.

\bibitem[{{AI@Meta}(2024)}]{meta-2024-llama3}
{AI@Meta}. 2024.
\newblock \href {https://github.com/meta-llama/llama3/blob/main/MODEL_CARD.md} {Llama 3 model card}.

\bibitem[{Appicharla et~al.(2025)Appicharla, Gain, Pal, and Ekbal}]{appicharla-2025-beyond}
Ramakrishna Appicharla, Baban Gain, Santanu Pal, and Asif Ekbal. 2025.
\newblock \href {https://doi.org/10.48550/arXiv.2506.07583} {Beyond the sentence: A survey on context-aware machine translation with large language models}.
\newblock \emph{arXiv preprint arXiv:2506.07583}.

\bibitem[{Cherry and Foster(2012)}]{cherry-foster-2012-batch}
Colin Cherry and George Foster. 2012.
\newblock \href {https://aclanthology.org/N12-1047/} {Batch tuning strategies for statistical machine translation}.
\newblock In \emph{Proceedings of NAACL-HLT}, pages 427--436.

\bibitem[{Christiano et~al.(2017)Christiano, Leike, Brown, Martic, Legg, and Amodei}]{christiano-2017-deep}
Paul Christiano, Jan Leike, Tom~B. Brown, Miljan Martic, Shane Legg, and Dario Amodei. 2017.
\newblock \href {https://proceedings.neurips.cc/paper_files/paper/2017/file/d5e2c0adad503c91f91df240d0cd4e49-Paper.pdf} {Deep reinforcement learning from human preferences}.
\newblock In \emph{Proceedings of NeurIPS}, pages 4302--4310.

\bibitem[{Cui et~al.(2025)Cui, Wang, Liu, Ke, Liu, and Bhat}]{cui-2025-crpo}
Guofeng Cui, Pichao Wang, Yang Liu, Zemian Ke, Zhu Liu, and Vimal Bhat. 2025.
\newblock \href {https://doi.org/10.18653/v1/2025.findings-acl.31} {{CRPO}: Confidence-reward driven preference optimization for machine translation}.
\newblock In \emph{Findings of ACL}, pages 560--574.

\bibitem[{Cui et~al.(2024)Cui, Du, Zhu, and Xiong}]{cui-2024-efficiently}
Menglong Cui, Jiangcun Du, Shaolin Zhu, and Deyi Xiong. 2024.
\newblock \href {https://aclanthology.org/2024.findings-acl.646/} {Efficiently exploring large language models for document-level machine translation with in-context learning}.
\newblock In \emph{Findings of ACL}, pages 10885--10897.

\bibitem[{Dong et~al.(2025)Dong, Lyu, Li, Wei, Zhang, Tao, and Yang}]{dong-2025-two}
Yichen Dong, Xinglin Lyu, Junhui Li, Daimeng Wei, Min Zhang, Shimin Tao, and Hao Yang. 2025.
\newblock \href {https://doi.org/10.18653/v1/2025.acl-long.726} {Two intermediate translations are better than one: Fine-tuning {LLM}s for document-level translation refinement}.
\newblock In \emph{Proceedings of ACL}, pages 14917--14933.

\bibitem[{Ferrando et~al.(2022)Ferrando, Gállego, Alastruey, Escolano, and Costa-jussà}]{ferrando-2022-towards}
Javier Ferrando, Gerard~I. Gállego, Belen Alastruey, Carlos Escolano, and Marta~R. Costa-jussà. 2022.
\newblock \href {https://aclanthology.org/2022.emnlp-main.599/} {Towards opening the black box of neural machine translation: Source and target interpretations of the transformer}.
\newblock In \emph{Proceedings of the EMNLP}, pages 8756--8769.

\bibitem[{Gao et~al.(2021)Gao, Yao, and Chen}]{gao-2021-simcse}
Tianyu Gao, Xingcheng Yao, and Danqi Chen. 2021.
\newblock \href {https://aclanthology.org/2021.emnlp-main.552} {{S}im{CSE}: Simple contrastive learning of sentence embeddings}.
\newblock In \emph{Proceedings of EMNLP}, pages 6894--6910.

\bibitem[{Guo et~al.(2025)Guo, Luo, Wei, Zhang, Li, Shang, Rao, Li, Yang, Wu, and Yang}]{guo-2025-doc}
Jiaxin Guo, Yuanchang Luo, Daimeng Wei, Ling Zhang, Zongyao Li, Hengchao Shang, Zhiqiang Rao, Shaojun Li, Jinlong Yang, Zhanglin Wu, and Hao Yang. 2025.
\newblock \href {https://arxiv.org/abs/2501.08523} {Doc-guided sent2sent++: A sent2sent++ agent with doc-guided memory for document-level machine translation}.
\newblock \emph{CoRR}, abs/2501.08523.

\bibitem[{Hopkins and May(2011)}]{hopkins-may-2011-tuning}
Mark Hopkins and Jonathan May. 2011.
\newblock \href {https://aclanthology.org/D11-1125/} {Tuning as ranking}.
\newblock In \emph{Proceedings of EMNLP}, pages 1352--1362.

\bibitem[{Hu et~al.(2022)Hu, Shen, Wallis, Allen-Zhu, Li, Wang, Wang, and Chen}]{hu-2022-lora}
Edward~J Hu, Yelong Shen, Phillip Wallis, Zeyuan Allen-Zhu, Yuanzhi Li, Shean Wang, Lu~Wang, and Weizhu Chen. 2022.
\newblock \href {https://openreview.net/forum?id=nZeVKeeFYf9} {Lora: Low-rank adaptation of large language models}.
\newblock In \emph{Proceedings of ICLR}.

\bibitem[{Hu et~al.(2025)Hu, Vamvas, and Sennrich}]{hu-2025-source}
Hanxu Hu, Jannis Vamvas, and Rico Sennrich. 2025.
\newblock \href {https://doi.org/10.18653/v1/2025.findings-emnlp.1289} {Source-primed multi-turn conversation helps large language models translate documents}.
\newblock In \emph{Findings of EMNLP}, pages 23702--23712.

\bibitem[{Jiang et~al.(2022)Jiang, Liu, Ma, Zhang, Yang, Huang, Sennrich, Cotterell, Sachan, and Zhou}]{jiang-2022-blonde}
Yuchen Jiang, Tianyu Liu, Shuming Ma, Dongdong Zhang, Jian Yang, Haoyang Huang, Rico Sennrich, Ryan Cotterell, Mrinmaya Sachan, and Ming Zhou. 2022.
\newblock \href {https://doi.org/10.18653/v1/2022.naacl-main.111} {{BlonDe}: An automatic evaluation metric for document-level machine translation}.
\newblock In \emph{Proceedings of the 2022 Conference of the North American Chapter of the Association for Computational Linguistics: Human Language Technologies}, pages 1550--1565.

\bibitem[{Karpinska and Iyyer(2023)}]{karpinska-iyyer-2023-large}
Marzena Karpinska and Mohit Iyyer. 2023.
\newblock \href {https://aclanthology.org/2023.wmt-1.41/} {Large language models effectively leverage document-level context for literary translation, but critical errors persist}.
\newblock In \emph{Proceedings of WMT}, pages 419--451.

\bibitem[{Li et~al.(2023)Li, Holtzman, Fried, Liang, Eisner, Hashimoto, Zettlemoyer, and Lewis}]{li-2023-contrastive}
Xiang~Lisa Li, Ari Holtzman, Daniel Fried, Percy Liang, Jason Eisner, Tatsunori Hashimoto, Luke Zettlemoyer, and Mike Lewis. 2023.
\newblock \href {https://aclanthology.org/2023.acl-long.687} {Contrastive decoding: Open-ended text generation as optimization}.
\newblock In \emph{Proceedings of ACL}, pages 12286--12312.

\bibitem[{Li et~al.(2024)Li, Li, Jiang, and Zhang}]{li-2024-enhancing}
Yachao Li, Junhui Li, Jing Jiang, and Min Zhang. 2024.
\newblock \href {https://arxiv.org/abs/2401.08088} {Enhancing document-level translation of large language model via translation mixed-instructions}.
\newblock \emph{CoRR}, abs/2401.08088.

\bibitem[{Li et~al.(2025)Li, Rao, Shang, Guo, Li, Wei, and Yang}]{li-2025-enhancing}
Zongyao Li, Zhiqiang Rao, Hengchao Shang, Jiaxin Guo, Shaojun Li, Daimeng Wei, and Hao Yang. 2025.
\newblock \href {https://aclanthology.org/2025.coling-main.591/} {Enhancing large language models for document-level translation post-editing using monolingual data}.
\newblock In \emph{Proceedings of COLING}, pages 8830--8840.

\bibitem[{Lyu et~al.(2024)Lyu, Li, Zhao, Zhang, Wei, Tao, Yang, and Zhang}]{lyu-2024-dempt}
Xinglin Lyu, Junhui Li, Yanqing Zhao, Min Zhang, Daimeng Wei, Shimin Tao, Hao Yang, and Min Zhang. 2024.
\newblock \href {https://doi.org/10.18653/v1/2024.emnlp-main.1131} {{D}e{MPT}: Decoding-enhanced multi-phase prompt tuning for making {LLM}s be better context-aware translators}.
\newblock In \emph{Proceedings of EMNLP}, pages 20280--20295.

\bibitem[{Maruf et~al.(2021)Maruf, Saleh, and Haffari}]{maruf-2021-survey}
Sameen Maruf, Fahimeh Saleh, and Gholamreza Haffari. 2021.
\newblock \href {https://doi.org/10.1145/3441691} {A survey on document-level neural machine translation: Methods and evaluation}.
\newblock \emph{ACM Computing Surveys}, 54(2).

\bibitem[{Meng et~al.(2024)Meng, Xia, and Chen}]{yu-2024-simpo}
Yu~Meng, Mengzhou Xia, and Danqi Chen. 2024.
\newblock \href {https://openreview.net/pdf?id=3Tzcot1LKb} {Simpo: Simple preference optimization with a reference-free reward}.
\newblock In \emph{Proceedings of NeurIPS}, pages 124198--124235.

\bibitem[{Mohammed and Niculae(2025)}]{mohammed-2025-context}
Wafaa Mohammed and Vlad Niculae. 2025.
\newblock \href {https://aclanthology.org/2025.mtsummit-1.10/} {Context-aware or context-insensitive? assessing {LLM}s' performance in document-level translation}.
\newblock In \emph{Proceedings of MT Summit}, pages 126--137.

\bibitem[{OpenAI(2025)}]{openai-2025-gpt5}
OpenAI. 2025.
\newblock \href {https://cdn.openai.com/gpt-5-system-card.pdf} {Gpt-5 system card}.

\bibitem[{Post(2018)}]{post-2018-call}
Matt Post. 2018.
\newblock \href {https://doi.org/10.18653/v1/W18-6319} {A call for clarity in reporting {BLEU} scores}.
\newblock In \emph{Proceedings of WMT}, pages 186--191.

\bibitem[{Rafailov et~al.(2023)Rafailov, Sharma, Mitchell, Ermon, Manning, and Finn}]{rafailov-2023-dpo}
Rafael Rafailov, Archit Sharma, Eric Mitchell, Stefano Ermon, Christopher~D. Manning, and Chelsea Finn. 2023.
\newblock \href {https://openreview.net/pdf?id=HPuSIXJaa9} {Direct preference optimization: Your language model is secretly a reward model}.
\newblock In \emph{Proceedings of NeurIPS}, pages 53728--53741.

\bibitem[{Ramos et~al.(2025)Ramos, Fernandes, Agrawal, and Martins}]{ramos-2025-docblocks}
Miguel~Moura Ramos, Patrick Fernandes, Sweta Agrawal, and Andr{\'e}~FT Martins. 2025.
\newblock \href {https://arxiv.org/abs/2504.12140} {Multilingual contextualization of large language models for document-level machine translation}.
\newblock \emph{arXiv preprint arXiv:2504.12140}.

\bibitem[{Rei et~al.(2022{\natexlab{a}})Rei, C.~de Souza, Alves, Zerva, Farinha, Glushkova, Lavie, Coheur, and Martins}]{rei-2022-comet}
Ricardo Rei, Jos{\'e}~G. C.~de Souza, Duarte Alves, Chrysoula Zerva, Ana~C Farinha, Taisiya Glushkova, Alon Lavie, Luisa Coheur, and Andr{\'e} F.~T. Martins. 2022{\natexlab{a}}.
\newblock \href {https://aclanthology.org/2022.wmt-1.52/} {{COMET}-22: Unbabel-{IST} 2022 submission for the metrics shared task}.
\newblock In \emph{Proceedings of WMT}, pages 578--585.

\bibitem[{Rei et~al.(2020)Rei, Stewart, Farinha, and Lavie}]{rei-2020-comet}
Ricardo Rei, Craig Stewart, Ana~C Farinha, and Alon Lavie. 2020.
\newblock \href {https://doi.org/10.18653/v1/2020.emnlp-main.213} {{COMET}: A neural framework for {MT} evaluation}.
\newblock In \emph{Proceedings of EMNLP}, pages 2685--2702, Online.

\bibitem[{Rei et~al.(2022{\natexlab{b}})Rei, Treviso, Guerreiro, Zerva, Farinha, Maroti, C.~de Souza, Glushkova, Alves, Coheur, Lavie, and Martins}]{rei-2022-cometkiwi}
Ricardo Rei, Marcos Treviso, Nuno~M. Guerreiro, Chrysoula Zerva, Ana~C Farinha, Christine Maroti, Jos{\'e}~G. C.~de Souza, Taisiya Glushkova, Duarte Alves, Luisa Coheur, Alon Lavie, and Andr{\'e} F.~T. Martins. 2022{\natexlab{b}}.
\newblock \href {https://aclanthology.org/2022.wmt-1.60} {{C}omet{K}iwi: {IST}-unbabel 2022 submission for the quality estimation shared task}.
\newblock In \emph{Proceedings of WMT}, pages 634--645.

\bibitem[{Stap et~al.(2024)Stap, Hasler, Byrne, Monz, and Tran}]{stap-2024-fine}
David Stap, Eva Hasler, Bill Byrne, Christof Monz, and Ke~Tran. 2024.
\newblock \href {https://aclanthology.org/2024.acl-long.336/} {The fine-tuning paradox: Boosting translation quality without sacrificing {LLM} abilities}.
\newblock In \emph{Proceedings of ACL}, pages 6189--6206.

\bibitem[{Sun et~al.(2025{\natexlab{a}})Sun, Gao, Zhang, Yang, and Wang}]{sun-2025-enhancing}
Haoxiang Sun, Ruize Gao, Pei Zhang, Baosong Yang, and Rui Wang. 2025{\natexlab{a}}.
\newblock \href {https://doi.org/10.18653/v1/2025.acl-long.1165} {Enhancing machine translation with self-supervised preference data}.
\newblock In \emph{Proceedings of ACL}, pages 23916--23934.

\bibitem[{Sun et~al.(2025{\natexlab{b}})Sun, Zhu, Chen, Xiao, Chen, and Shen}]{sun-2025-fine}
Yirong Sun, Dawei Zhu, Yanjun Chen, Erjia Xiao, Xinghao Chen, and Xiaoyu Shen. 2025{\natexlab{b}}.
\newblock \href {https://doi.org/10.18653/v1/2025.naacl-srw.1} {Fine-grained and multi-dimensional metrics for document-level machine translation}.
\newblock In \emph{Proceedings of NAACL-HLT: Student Research Workshop}, pages 1--17.

\bibitem[{Tiedemann and Scherrer(2017)}]{tiedemann-2017-neural}
J{\"o}rg Tiedemann and Yves Scherrer. 2017.
\newblock \href {https://aclanthology.org/W17-4811/} {Neural machine translation with extended context}.
\newblock In \emph{Proceedings of the Third Workshop on Discourse in Machine Translation}, pages 82--92.

\bibitem[{Uhlig et~al.(2025)Uhlig, Wuebker, Reinauer, and Denero}]{uhlig-2025-dqo}
Kaden Uhlig, Joern Wuebker, Raphael Reinauer, and John Denero. 2025.
\newblock \href {https://doi.org/10.18653/v1/2025.wmt-1.2} {Cross-lingual human-preference alignment for neural machine translation with direct quality optimization}.
\newblock In \emph{Proceedings of WMT}, pages 31--51.

\bibitem[{Vernikos et~al.(2022)Vernikos, Thompson, Mathur, and Federico}]{vernikos-2022-embarrassingly}
Giorgos Vernikos, Brian Thompson, Prashant Mathur, and Marcello Federico. 2022.
\newblock \href {https://aclanthology.org/2022.wmt-1.6/} {Embarrassingly easy document-level {MT} metrics: How to convert any pretrained metric into a document-level metric}.
\newblock In \emph{Proceedings of WMT}, pages 118--128.

\bibitem[{Voita et~al.(2018)Voita, Serdyukov, Sennrich, and Titov}]{voita-2018-context}
Elena Voita, Pavel Serdyukov, Rico Sennrich, and Ivan Titov. 2018.
\newblock \href {https://doi.org/10.18653/v1/P18-1117} {Context-aware neural machine translation learns anaphora resolution}.
\newblock In \emph{Proceedings of ACL}, pages 1264--1274.

\bibitem[{Wang et~al.(2026{\natexlab{a}})Wang, Xu, Liu, Liu, Zhao, Zeng, Shao, Dong, Wu, Zhou, Dong, Zeng, Wang, and Luo}]{wang-2026-m2po}
Hao Wang, Linlong Xu, Heng Liu, Yangyang Liu, Xiaohu Zhao, Bo~Zeng, Liangying Shao, Yichen Dong, Xinwei Wu, Jiang Zhou, Tianyu Dong, Xiangxiang Zeng, Longyue Wang, and Weihua Luo. 2026{\natexlab{a}}.
\newblock \href {https://doi.org/10.18653/v1/2026.acl-long.469} {{M}$^2${PO}: Multi-perspective multi-pair preference optimization for machine translation}.
\newblock In \emph{Proceedings of ACL}, pages 10315--10336.

\bibitem[{Wang et~al.(2023)Wang, Lyu, Ji, Zhang, Yu, Shi, and Tu}]{wang-2023-document-level}
Longyue Wang, Chenyang Lyu, Tianbo Ji, Zhirui Zhang, Dian Yu, Shuming Shi, and Zhaopeng Tu. 2023.
\newblock \href {https://aclanthology.org/2023.emnlp-main.1036} {Document-level machine translation with large language models}.
\newblock In \emph{Proceedings of EMNLP}, pages 16646--16661.

\bibitem[{Wang et~al.(2026{\natexlab{b}})Wang, Liu, Wong, Li, Jiang, Zhang, Tao, Wei, and Zhang}]{wang-2026-loong}
Yutong Wang, Xuebo Liu, Derek~F. Wong, Zhilin Li, Rongqing Jiang, Min Zhang, Shimin Tao, Daimeng Wei, and Min Zhang. 2026{\natexlab{b}}.
\newblock \href {https://arxiv.org/abs/2605.30274} {{Loong}: A human-like long document translation agent with observe-and-act adaptive context selection}.
\newblock \emph{arXiv preprint arXiv:2605.30274}.

\bibitem[{Wang et~al.(2025)Wang, Zeng, Liu, Wong, Meng, Zhou, and Zhang}]{wang-2025-delta}
Yutong Wang, Jiali Zeng, Xuebo Liu, Derek~F. Wong, Fandong Meng, Jie Zhou, and Min Zhang. 2025.
\newblock \href {https://openreview.net/forum?id=hoYFLRNbhc} {Delta: An online document-level translation agent based on multi-level memory}.
\newblock In \emph{Proceedings of ICLR}.

\bibitem[{Wu et~al.(2024{\natexlab{a}})Wu, Vu, Qu, Foster, and Haffari}]{wu-2024-adapting}
Minghao Wu, Thuy-Trang Vu, Lizhen Qu, George Foster, and Gholamreza Haffari. 2024{\natexlab{a}}.
\newblock \href {https://arxiv.org/abs/2401.06468} {Adapting large language models for document-level machine translation}.
\newblock \emph{CoRR}, abs/2401.06468.

\bibitem[{Wu et~al.(2024{\natexlab{b}})Wu, Nagata, Miao, and Tsuruoka}]{wu-2024-word}
Qiyu Wu, Masaaki Nagata, Zhongtao Miao, and Yoshimasa Tsuruoka. 2024{\natexlab{b}}.
\newblock \href {https://doi.org/10.18653/v1/2024.emnlp-main.188} {Word alignment as preference for machine translation}.
\newblock In \emph{Proceedings of EMNLP}, pages 3223--3239.

\bibitem[{Xu et~al.(2024)Xu, Sharaf, Chen, Tan, Shen, Van~Durme, Murray, and Kim}]{xu-2024-cpo}
Haoran Xu, Amr Sharaf, Yunmo Chen, Weiting Tan, Lingfeng Shen, Benjamin Van~Durme, Kenton Murray, and Young~Jin Kim. 2024.
\newblock \href {https://proceedings.mlr.press/v235/xu24t.html} {Contrastive preference optimization: pushing the boundaries of llm performance in machine translation}.
\newblock In \emph{Proceedings of ICML}, pages 55204--55224.

\bibitem[{Yang et~al.(2025)Yang, Li, Yang, Zhang, Hui, and Others}]{yang-2025-qwen3}
An~Yang, Anfeng Li, Baosong Yang, Beichen Zhang, Binyuan Hui, and Others. 2025.
\newblock \href {https://arxiv.org/abs/2505.09388} {Qwen3 technical report}.
\newblock \emph{arXiv preprint arXiv:2505.09388}.

\bibitem[{Yang et~al.(2024)Yang, Chen, Lin, and Byrne}]{yang-2024-direct}
Guangyu Yang, Jinghong Chen, Weizhe Lin, and Bill Byrne. 2024.
\newblock \href {https://doi.org/10.18653/v1/2024.naacl-short.34} {Direct preference optimization for neural machine translation with minimum {B}ayes risk decoding}.
\newblock In \emph{Proceedings of NAACL-HLT: Short Paper}, pages 391--398.

\end{thebibliography}

\newpage

\appendix

\section{Preference Pair Selection}\label{apx:pair_selection}

We construct the sentence-level preference set $\mathcal{P}_{\text{s}}$ by filtering the preferred–dispreferred translation pairs $(y_{\text{s}}^{+}, y_{\text{s}}^{-})$ obtained in Section~\ref{sec:two_tasks}. A pair is retained only if it satisfies the following criteria: 1) both translations contain between 6 and 100 words; 2) both translations achieve quality scores higher than 30; and 3) the score margin between the two translations lies in the range $[0.2, 10]$.

These constraints filter out degenerate, low-quality, or weakly distinguishable pairs, ensuring that retained pairs provide reliable and informative preference signals. The same strategy is applied to construct the context-aware preference set $\mathcal{P}_{\text{c}}$.

Similarly, all preference pairs in the cross-preference set $\mathcal{P}_{\text{cr}}$ are required to satisfy the same selection criteria.

\section{Dataset}\label{apx:dataset}
\noindent\textbf{Data splits.}
For News Commentary v18.1, we follow the document-level train/validation/test
split of \citet{dong-2025-two} and \citet{li-2024-enhancing}. For IWSLT2017,
we adopt the original released splits, using the official development set
(\texttt{dev2010}) as the validation set and the union of
\texttt{tst2010}--\texttt{tst2015} as the test set. In both domains, the
training split is used for the cold-start fine-tuning stage and the validation
split is the sole data source for preference selection, so no test-side
information is involved in constructing preference pairs. For fair comparison,
Sent+Ctx$_{\text{tuned+}}$, Sent-Level$_{\text{CPO}}$, and
Ctx-Aware$_{\text{CPO}}$ draw on the same splits (Section~\ref{sec:setting}).

\noindent\textbf{Data statistics.}
Table~\ref{tab:stat} shows the statistics of the test sets of the fifteen translation tasks, and the dataset used for CPL preference selection. These tasks cover two domains, News and IWSLT, and include both the En$\Rightarrow$X direction (En$\Rightarrow$De/Es/Fr/It/Nl/Ru) and the X$\Rightarrow$En direction (Es/Fr/Nl/Ru/Zh$\Rightarrow$En) on News, as well as IWSLT (En$\Rightarrow$De, En$\Rightarrow$Fr, En$\Rightarrow$Zh, Zh$\Rightarrow$En).

Table~\ref{tab:stat_cpl} shows the statistics of the preference datasets used in these tasks. Specifically, $\mathcal{P}_{\text{s}}$, $\mathcal{P}_{\text{c}}$, and $\mathcal{P}_{\text{cr}}$ (as defined in Section~\ref{sec:method}) denote the numbers of intra-condition sentence-level, intra-condition context-level, and cross-condition preference pairs, respectively, where $\mathcal{P}_{\text{cr}}$ includes both $(y_{\text{w}}^{+},y_{\text{l}}^{+})$ and $(y_{\text{w}}^{+},y_{\text{l}}^{-})$.

\begin{table}[t]
\centering
\small
\setlength{\tabcolsep}{4pt}
\begin{tabular}{llrrrr}
\toprule
\multirow{2}{*}{Data} & \multirow{2}{*}{Task} & \multicolumn{2}{c}{Test} & \multicolumn{2}{c}{CPL} \\
\cmidrule(lr){3-4}\cmidrule(lr){5-6}
 & & \#Doc & \#Sent & \#Doc & \#Sent \\
\midrule
\multirow{6}{*}{\shortstack[l]{News\\\scriptsize(En$\Rightarrow$X)}}
 & En $\Rightarrow$ De & 150 & 5,967 & 150 & 5,628 \\
 & En $\Rightarrow$ Es & 150 & 5,819 & 150 & 5,782 \\
 & En $\Rightarrow$ Fr & 150 & 5,795 & 150 & 5,890 \\
 & En $\Rightarrow$ It & 150 & 5,749 & 150 & 5,948 \\
 & En $\Rightarrow$ Nl & 151 & 5,992 & 151 & 5,925 \\
 & En $\Rightarrow$ Ru & 150 & 5,619 & 150 & 5,691 \\
\midrule
\multirow{5}{*}{\shortstack[l]{News\\\scriptsize(X$\Rightarrow$En)}}
 & Es $\Rightarrow$ En & 150 & 5,819 & 150 & 5,782 \\
 & Fr $\Rightarrow$ En & 150 & 5,795 & 150 & 5,890 \\
 & Nl $\Rightarrow$ En & 151 & 5,992 & 151 & 5,925 \\
 & Ru $\Rightarrow$ En & 150 & 5,619 & 150 & 5,691 \\
 & Zh $\Rightarrow$ En & 150 & 5,908 & 150 & 6,061 \\
\midrule
\multirow{4}{*}{IWSLT}
 & En $\Rightarrow$ De & 85 & 8,079 & 8 & 888 \\
 & En $\Rightarrow$ Fr & 89 & 8,597 & 8 & 890 \\
 & En $\Rightarrow$ Zh & 89 & 8,549 & 8 & 879 \\
 & Zh $\Rightarrow$ En & 89 & 8,549 & 8 & 879 \\
\bottomrule
\end{tabular}
\caption{Statistics of the test sets and of the data used for CPL preference selection, across the fifteen translation tasks in the News and IWSLT domains. \#Doc and \#Sent denote the numbers of documents and sentences.}
\label{tab:stat}
\end{table}

\begin{table}[t]
\centering
\small
\setlength{\tabcolsep}{6pt}
\begin{tabular}{llrrr}
\toprule
Data & Task & $\mathcal{P}_{\text{s}}$ & $\mathcal{P}_{\text{c}}$ & $\mathcal{P}_{\text{cr}}$ \\
\midrule
\multirow{6}{*}{\shortstack[l]{News\\\scriptsize(En$\Rightarrow$X)}}
 & En $\Rightarrow$ De & 4,143 & 4,277 & 8,086 \\
 & En $\Rightarrow$ Es & 4,963 & 5,051 & 8,736 \\
 & En $\Rightarrow$ Fr & 4,768 & 4,877 & 8,829 \\
 & En $\Rightarrow$ It & 4,923 & 5,024 & 8,877 \\
 & En $\Rightarrow$ Nl & 3,904 & 4,072 & 8,056 \\
 & En $\Rightarrow$ Ru & 4,016 & 4,116 & 7,841 \\
\midrule
\multirow{5}{*}{\shortstack[l]{News\\\scriptsize(X$\Rightarrow$En)}}
 & Es $\Rightarrow$ En & 5,224 & 5,233 & 8,947 \\
 & Fr $\Rightarrow$ En & 5,343 & 5,376 & 9,289 \\
 & Nl $\Rightarrow$ En & 5,367 & 5,440 & 9,214 \\
 & Ru $\Rightarrow$ En & 4,903 & 4,973 & 8,688 \\
 & Zh $\Rightarrow$ En & 5,496 & 5,525 & 9,716 \\
\midrule
\multirow{4}{*}{IWSLT}
 & En $\Rightarrow$ De & 459 & 471 & 986 \\
 & En $\Rightarrow$ Fr & 495 & 477 & 1,003 \\
 & En $\Rightarrow$ Zh & 375 & 418 & 867 \\
 & Zh $\Rightarrow$ En & 572 & 579 & 1,087 \\
\bottomrule
\end{tabular}
\caption{Statistics of the preference pairs in cross-condition preference learning, across the fifteen translation tasks. $\mathcal{P}_{\text{s}}$, $\mathcal{P}_{\text{c}}$, and $\mathcal{P}_{\text{cr}}$ denote the numbers of intra-condition sentence-level, intra-condition context-level, and cross-condition preference pairs, respectively.}
\label{tab:stat_cpl}
\end{table}

\section{SFT and CPL Settings}
\label{apx:settings}
During SFT, we fine-tune the model using the LoRA approach~\cite{hu-2022-lora}. The LoRA rank is set to 16. The batch size is 256, and the learning rate is fixed at $5 \times 10^{-5}$. We train the model for a single epoch.

During CPL, we also adopt LoRA-based fine-tuning with the LoRA rank set to 16. The batch size is 128, and the learning rate is set to $5 \times 10^{-5}$. The model is trained for 2 epochs. The $\beta$ parameter is set to 0.1, following \citet{rafailov-2023-dpo} and \citet{xu-2024-cpo}.

\section{Experimental Results in d-COMET and s-BLEU}
\label{apx:more_results}

Table~\ref{tab:d-comet} shows the performance in document-level COMET~\cite{vernikos-2022-embarrassingly} using the reference-based \texttt{wmt22-comet-da}~\cite{rei-2022-comet}. Table~\ref{tab:bleu} shows the performance in sentence-level SacreBLEU~\cite{post-2018-call}.

\begin{table*}[!t]
\centering
\small
\resizebox{\textwidth}{!}{
\begin{NiceTabular}[color-inside]{l|l|rcrcrcrcrcrc|rc}
\toprule
\Block[l]{2-1}{\textbf{\#}} & \Block[l]{2-1}{\textbf{Model}} & \Block[c]{1-2}{\textbf{En $\Rightarrow$ De}} &  & \Block[c]{1-2}{\textbf{En $\Rightarrow$ Es}} & & \Block[c]{1-2}{\textbf{En $\Rightarrow$ Fr}} & & \Block[c]{1-2}{\textbf{En $\Rightarrow$ It}} & & \Block[c]{1-2}{\textbf{En $\Rightarrow$ Nl}} && \Block[c]{1-2}{\textbf{En $\Rightarrow$ Ru}} & & \Block[c]{1-2}{\textbf{\textit{Average}}} &\\

\cmidrule(lr){3-4}\cmidrule(lr){5-6}\cmidrule(lr){7-8}\cmidrule(lr){9-10}\cmidrule(lr){11-12}\cmidrule(lr){13-14}\cmidrule(lr){15-16}
& & \textit{Sent} & \textit{CTX} & \textit{Sent} & \textit{CTX}  & \textit{Sent} & \textit{CTX}  & \textit{Sent} & \textit{CTX} & \textit{Sent} & \textit{CTX} & \textit{Sent} & \textit{CTX} & \textit{Sent} & \textit{CTX} \\

\rowcolor[gray]{0.85}
\Block[c]{1-16}{\texttt{Qwen3-4B}}\\
\cellcolor{sentlevel}1 & \cellcolor{sentlevel}Sent-Level & 79.00 & - & 81.29 & - & 79.50 & - & 83.26 & - & 79.35 & - & 79.23 & - & 80.27 & - \\
\cellcolor{sentlevel}2 & \cellcolor{sentlevel}Sent-Level$_\text{tuned}$ & 81.55 & - & 84.25 & - & 81.74 & - & 84.81 & - & 81.94 & - & 81.02 & - & 82.55 & - \\
\cellcolor{sentlevel}3 & \cellcolor{sentlevel}Sent-Level$_\text{CPO}$ & 82.40 & - & 84.90 & - & 82.37 & - & 85.57 & - & 82.98 & - & 82.73 & - & 83.49 & - \\
\cellcolor{ctxlevel}4 & \cellcolor{ctxlevel}Ctx-Aware & - & 80.26 & - & 84.46 & - & 81.59 & - & 84.38 & - & 80.29 & - & 80.22 & - & 81.87 \\
\cellcolor{ctxlevel}5 & \cellcolor{ctxlevel}Ctx-Aware$_\text{tuned}$ & - & 81.52 & - & 84.26 & - & 81.84 & - & 84.85 & - & 81.95 & - & 81.38 & - & 82.63 \\
\cellcolor{ctxlevel}6 & \cellcolor{ctxlevel}Ctx-Aware$_\text{CPO}$ & - & 82.74 & - & 85.03 & - & 82.60 & - & 85.74 & - & 83.15 & - & 82.83 & - & 83.68 \\
\cellcolor{sentctxlevel}7 & \cellcolor{sentctxlevel}Sent+Ctx$_{\text{tuned}}$ & 81.78 & 82.00 & 84.44 & 84.46 & 81.70 & 81.83 & 85.04 & 85.20 & 82.24 & 82.20 & 81.53 & 81.43 & 82.79 & 82.85 \\
\cellcolor{sentctxlevel}8 & \cellcolor{sentctxlevel}Sent+Ctx$_{\text{tuned+}}$ & 81.72 & 81.85 & 84.49 & 84.55 & 81.87 & 81.96 & 85.29 & 85.37 & 82.43 & 82.53 & 81.31 & 81.26 & 82.85 & 82.92 \\
\cellcolor{our}9 & \cellcolor{our}CPL (Ours) & \cellcolor{darkred!80}83.23 & \cellcolor{darkred!80}83.25 & \cellcolor{darkred!80}85.22 & \cellcolor{darkred!80}85.38 & \cellcolor{darkred!80}83.02 & \cellcolor{darkred!80}83.08 & \cellcolor{darkred!80}86.00 & \cellcolor{darkred!80}86.04 & \cellcolor{darkred!80}83.82 & \cellcolor{darkred!80}83.84 & \cellcolor{darkred!80}83.01 & \cellcolor{darkred!80}83.09 & \cellcolor{darkred!80}84.05 & \cellcolor{darkred!80}84.11 \\
\cellcolor{our}10 & \cellcolor{our}~~ - IntraLoss & 82.82 & 82.83 & \cellcolor{lightred!80}85.10 & 85.11 & \cellcolor{lightred!80}82.76 & 82.78 & \cellcolor{lightred!80}85.81 & 85.91 & 83.39 & 83.48 & 82.64 & 82.53 & 83.75 & 83.77 \\
\cellcolor{our}11 & \cellcolor{our}~~ - CrossLoss & \cellcolor{lightred!80}82.86 & \cellcolor{lightred!80}83.02 & 85.07 & \cellcolor{lightred!80}85.24 & \cellcolor{lightred!80}82.76 & \cellcolor{lightred!80}82.95 & 85.76 & \cellcolor{lightred!80}85.93 & \cellcolor{lightred!80}83.67 & \cellcolor{lightred!80}83.72 & \cellcolor{lightred!80}82.92 & \cellcolor{lightred!80}83.01 & \cellcolor{lightred!80}83.84 & \cellcolor{lightred!80}83.98 \\

\rowcolor[gray]{0.85}
\Block[c]{1-16}{\texttt{Qwen3-8B}}\\
\cellcolor{sentlevel}1 & \cellcolor{sentlevel}Sent-Level & 82.15 & - & 85.01 & -  & 82.38 & -  & 85.51 & -  & 83.08 & -  & 82.04  & - & 83.36 & -  \\
\cellcolor{sentlevel}2 & \cellcolor{sentlevel}Sent-Level$_\text{tuned}$  & 83.37 & - & 85.17 & -  & 82.99 & -  & 86.06 & -  & 84.37 & -  & 82.88  & - & 84.14 & -  \\
\cellcolor{sentlevel}3 & \cellcolor{sentlevel}Sent-Level$_\text{CPO}$ & 83.65 & - & 85.69 & -  & 83.41 & -  & 86.44 & -  & 85.07 & -  & 83.91 & - & 84.69 & -  \\
\cellcolor{ctxlevel}4 & \cellcolor{ctxlevel}Ctx-Aware & - & 82.29  & -  & 85.15  & -  & 82.59 & -  & 85.56 & -  & 83.40 & - & 81.99  & - & 83.50 \\
\cellcolor{ctxlevel}5 & \cellcolor{ctxlevel}Ctx-Aware$_\text{tuned}$ & - & 83.43 & -  & 85.25 & -  & 83.04 & -  & 86.00 & -  & 84.48  & - & 82.97  & - & 84.20 \\
\cellcolor{ctxlevel}6 & \cellcolor{ctxlevel}Ctx-Aware$_\text{CPO}$  & - & 83.95 & -  & 85.79 & -  & 83.56 & -  & 86.53 & -  & 85.03 & - & 83.97 & - & 84.81 \\
\cellcolor{sentctxlevel}7 & \cellcolor{sentctxlevel}Sent+Ctx$_{\text{tuned}}$ & 83.39 & 83.58  & 85.27 & 85.35 & 82.89 & 83.03 & 86.15 & 86.23 & 84.69 & 84.74 & 83.03 & 82.97 & 84.24 & 84.32 \\
\cellcolor{sentctxlevel}8 & \cellcolor{sentctxlevel}Sent+Ctx$_{\text{tuned+}}$ & 83.49 & 83.68 & 85.24 & 85.41 & 82.98 & 83.13 & 86.25 & 86.25 & 84.72 & 84.91 & 82.90 & 82.71 & 84.26 & 84.35 \\
\cellcolor{our}9 & \cellcolor{our}CPL (Ours)& \cellcolor{darkred!80}84.63 & \cellcolor{darkred!80}84.69 & \cellcolor{darkred!80}86.00 & \cellcolor{darkred!80}86.05 & \cellcolor{darkred!80}83.82 & \cellcolor{darkred!80}83.92 & \cellcolor{lightred!80}86.78 & \cellcolor{darkred!80}87.05 & \cellcolor{darkred!80}85.69 & \cellcolor{darkred!80}85.84 & \cellcolor{darkred!80}84.32 & \cellcolor{darkred!80}84.36 & \cellcolor{darkred!80}85.21 & \cellcolor{darkred!80}85.32 \\
\cellcolor{our}10 & \cellcolor{our}~~ - IntraLoss & 84.23 & 84.32 & \cellcolor{lightred!80}85.85 & 85.88 & 83.64 & 83.64 & 86.69 & 86.69 & 85.32 & 85.33 & \cellcolor{lightred!80}83.98 & 83.87 & 84.95 & 84.95 \\
\cellcolor{our}11 & \cellcolor{our}~~ - CrossLoss & \cellcolor{lightred!80}84.29 & \cellcolor{lightred!80}84.44 & 85.83 & \cellcolor{lightred!80}85.91 & \cellcolor{lightred!80}83.68 & \cellcolor{lightred!80}83.85 & \cellcolor{darkred!80}86.84 & \cellcolor{lightred!80}86.81 & \cellcolor{lightred!80}85.42 & \cellcolor{lightred!80}85.45 & \cellcolor{lightred!80}83.98 & \cellcolor{lightred!80}84.02 & \cellcolor{lightred!80}85.01 & \cellcolor{lightred!80}85.08 \\

\rowcolor[gray]{0.85}
\Block[c]{1-16}{\texttt{Llama-3-8B-Instruct}}\\
\cellcolor{sentlevel}1 & \cellcolor{sentlevel}Sent-Level & 78.32 & - & 80.56 & -  & 79.22 & -  & 81.46 & -  & 81.20 & -  &	77.15  & - & 79.65 & -  \\
\cellcolor{sentlevel}2 & \cellcolor{sentlevel}Sent-Level$_\text{tuned}$  & 83.26 & - & 85.07 & -  & 82.54 & -  & 85.93 & -  & 85.63 & -  & 82.21  & - & 84.11 & -  \\
\cellcolor{sentlevel}3 & \cellcolor{sentlevel}Sent-Level$_\text{CPO}$ & 83.86 & - & 85.63 & -  & 83.15 & -  & 86.58 & -  & 86.12 & -  & 83.27 & - & 84.77 & -  \\
\cellcolor{ctxlevel}4 & \cellcolor{ctxlevel}Ctx-Aware & - & 44.93  & -  & 50.38  & -  & 44.08 & -  & 51.81 & -  & 46.62 & - & 45.98  & - & 47.30 \\
\cellcolor{ctxlevel}5 & \cellcolor{ctxlevel}Ctx-Aware$_\text{tuned}$ & - & 83.39 & -  & 85.20 & -  & 82.80 & -  & 86.02 & -  & 85.73  & - & 82.34  & - & 84.25 \\
\cellcolor{ctxlevel}6 & \cellcolor{ctxlevel}Ctx-Aware$_\text{CPO}$  & - & 84.14 & -  & 85.67 & -  & 83.47 & -  & 86.61 & -  & 86.20 & - & 83.38 & - & 84.91 \\
\cellcolor{sentctxlevel}7 & \cellcolor{sentctxlevel}Sent+Ctx$_{\text{tuned}}$ & 83.69 & 83.84  & 85.34 & 85.38 & 82.95 & 83.05	& 86.29 & 86.29 	& 86.14 & 86.20	& 82.77 & 82.64 & 84.53 & 84.57 \\
\cellcolor{sentctxlevel}8 & \cellcolor{sentctxlevel}Sent+Ctx$_{\text{tuned+}}$ & 83.67 & 83.81 & 85.39 & 85.43 & 83.12 & 83.20 & 86.42 & 86.36 & 86.30 & 86.34 & 82.70 & 82.73 & 84.60 & 84.65 \\
\cellcolor{our}9 & \cellcolor{our}CPL (Ours)& \cellcolor{darkred!80}84.44 & \cellcolor{darkred!80}84.59 &\cellcolor{darkred!80} 85.98 &\cellcolor{darkred!80} 86.04 &\cellcolor{darkred!80} 83.84 &\cellcolor{darkred!80} 83.95 & \cellcolor{darkred!80}86.83 &\cellcolor{darkred!80} 86.89 &\cellcolor{darkred!80} 86.49 &\cellcolor{darkred!80} 86.58 & \cellcolor{darkred!80}83.99 &\cellcolor{darkred!80} 83.99 &\cellcolor{darkred!80} 85.26 &\cellcolor{darkred!80} 85.34 \\
\cellcolor{our}10 & \cellcolor{our}~~ - IntraLoss & 
84.41 & 84.45 & \cellcolor{lightred!80}85.79 & \cellcolor{lightred!80}85.84 & 83.62 & \cellcolor{lightred!80}83.76 &
86.63 & \cellcolor{lightred!80}86.76 & 86.34 & 86.37 & 83.65 & 83.54 & 85.07 & 85.12 \\
\cellcolor{our}11 & \cellcolor{our}~~ - CrossLoss & 
\cellcolor{lightred!80}84.42 & \cellcolor{lightred!80}84.56 &
85.65 & 85.71 &
\cellcolor{lightred!80}83.69 & 83.75 &
\cellcolor{lightred!80}86.68 & 86.73 &
\cellcolor{lightred!80}86.48 & \cellcolor{lightred!80}86.55 &
\cellcolor{lightred!80}83.89 & \cellcolor{lightred!80}83.98 &
\cellcolor{lightred!80}85.14 & \cellcolor{lightred!80}85.21 \\

\bottomrule
\end{NiceTabular}
}
\caption{Results of different systems on document-level COMET metric. Best scores are shown in \colorbox{darkred!80}{dark red} and second-best scores are in \colorbox{lightred!80}{light red}. \textit{Sent} and \textit{CTX} denote sentence-level and context-aware inputs, respectively.}
\label{tab:d-comet}
\end{table*}


\begin{table*}[!t]
\centering
\small
\resizebox{\textwidth}{!}{
\begin{NiceTabular}[color-inside]{l|l|rcrcrcrcrcrc|rc}
\toprule
\Block[l]{2-1}{\textbf{\#}} & \Block[l]{2-1}{\textbf{Model}} & \Block[c]{1-2}{\textbf{En $\Rightarrow$ De}} &  & \Block[c]{1-2}{\textbf{En $\Rightarrow$ Es}} & & \Block[c]{1-2}{\textbf{En $\Rightarrow$ Fr}} & & \Block[c]{1-2}{\textbf{En $\Rightarrow$ It}} & & \Block[c]{1-2}{\textbf{En $\Rightarrow$ Nl}} && \Block[c]{1-2}{\textbf{En $\Rightarrow$ Ru}} & & \Block[c]{1-2}{\textbf{\textit{Average}}} &\\

\cmidrule(lr){3-4}\cmidrule(lr){5-6}\cmidrule(lr){7-8}\cmidrule(lr){9-10}\cmidrule(lr){11-12}\cmidrule(lr){13-14}\cmidrule(lr){15-16}
& & \textit{Sent} & \textit{CTX} & \textit{Sent} & \textit{CTX}  & \textit{Sent} & \textit{CTX}  & \textit{Sent} & \textit{CTX} & \textit{Sent} & \textit{CTX} & \textit{Sent} & \textit{CTX} & \textit{Sent} & \textit{CTX} \\

\rowcolor[gray]{0.85}
\Block[c]{1-16}{\texttt{Qwen3-4B}}\\

\cellcolor{sentlevel}1 & \cellcolor{sentlevel}Sent-Level & 12.64 & - & 28.95 & - & 22.16 & - & 22.75 & - & 14.78 & - & 14.38 & - & 19.28 & - \\
\cellcolor{sentlevel}2 & \cellcolor{sentlevel}Sent-Level$_\text{tuned}$ & 24.92 & - & 40.05 & - & 31.94 & - & 35.38 & - & 24.38 & - & 22.20 & - & 29.81 & - \\
\cellcolor{sentlevel}3 & \cellcolor{sentlevel}Sent-Level$_\text{CPO}$ & 28.99 & - & 41.73 & - & \cellcolor{lightred!80}35.16 & - & 36.25 & - & 28.61 & - & 26.49 & - & 32.87 & - \\
\cellcolor{ctxlevel}4 & \cellcolor{ctxlevel}Ctx-Aware & - & 11.09 & - & 28.98 & - & 21.21 & - & 21.35 & - & 12.72 & - & 13.05 & - & 18.07 \\
\cellcolor{ctxlevel}5 & \cellcolor{ctxlevel}Ctx-Aware$_\text{tuned}$ & - & 26.73 & - & 40.02 & - & 32.79 & - & 35.20 & - & 24.46 & - & 23.26 & - & 30.41 \\
\cellcolor{ctxlevel}6 & \cellcolor{ctxlevel}Ctx-Aware$_\text{CPO}$ & - & 27.71 & - & 41.29 & - & 34.94 & - & 35.14 & - & 30.30 & - & 24.99 & - & 32.40 \\
\cellcolor{sentctxlevel}7 & \cellcolor{sentctxlevel}Sent+Ctx$_{\text{tuned}}$ & 28.11 & 28.95 & 41.33 & 39.42 & 33.54 & 33.38 & \cellcolor{lightred!80}36.65 & 36.56 & 27.51 & 28.32 & 24.89 & 25.52 & 32.01 & 32.03 \\
\cellcolor{sentctxlevel}8 & \cellcolor{sentctxlevel}Sent+Ctx$_{\text{tuned+}}$ & 28.42 & 29.15 & 41.47 & 40.74 & 32.84 & 33.46 & \cellcolor{darkred!80}37.26 & \cellcolor{darkred!80}37.52 & 29.74 & 28.17 & 24.79 & 25.19 & 32.42 & 32.37 \\
\cellcolor{our}9 & \cellcolor{our}CPL (Ours) & \cellcolor{lightred!80}29.38 & 29.46 & \cellcolor{darkred!80}42.03 & \cellcolor{darkred!80}41.99 & \cellcolor{darkred!80}35.50 & \cellcolor{darkred!80}35.63 & 35.87 & 35.32 & \cellcolor{darkred!80}33.16 & \cellcolor{darkred!80}33.24 & \cellcolor{darkred!80}27.38 & \cellcolor{darkred!80}27.39 & \cellcolor{lightred!80}33.89 & \cellcolor{lightred!80}33.84 \\
\cellcolor{our}10 & \cellcolor{our}~~ - IntraLoss & 29.10 & \cellcolor{darkred!80}30.03 & 41.64 & 41.75 & 35.11 & \cellcolor{lightred!80}35.27 & 36.00 & \cellcolor{lightred!80}36.69 & 32.71 & \cellcolor{lightred!80}32.80 & 26.30 & \cellcolor{lightred!80}27.18 & 33.48 & \cellcolor{darkred!80}33.95 \\
\cellcolor{our}11 & \cellcolor{our}~~ - CrossLoss & \cellcolor{darkred!80}30.56 & \cellcolor{lightred!80}29.98 & \cellcolor{lightred!80}41.87 & \cellcolor{lightred!80}41.87 & 34.89 & 34.97 & 36.37 & 36.01 & \cellcolor{lightred!80}32.91 & 32.21 & \cellcolor{lightred!80}26.78 & 26.80 & \cellcolor{darkred!80}33.90 & 33.64 \\

\rowcolor[gray]{0.85}
\Block[c]{1-16}{\texttt{Qwen3-8B}}\\\cellcolor{sentlevel}1 & \cellcolor{sentlevel}Sent-Level & 16.95 & - & 35.96 & -  & 25.91 & -  & 27.32 & -  & 18.39 & -  & 18.41  & - & 23.82 & -  \\
\cellcolor{sentlevel}2 & \cellcolor{sentlevel}Sent-Level$_\text{tuned}$  & 31.72 & - & 42.65 & -  & 35.48 & -  & 38.59 & -  & 34.66 & -  & 28.66  & - & 35.29 & -  \\
\cellcolor{sentlevel}3 & \cellcolor{sentlevel}Sent-Level$_\text{CPO}$ & 33.47 & - & 43.79 & -  & 37.40 & -  & 38.90 & -  & 35.01 & -  & \cellcolor{lightred!80}30.12 & - & 36.45 & -  \\
\cellcolor{ctxlevel}4 & \cellcolor{ctxlevel}Ctx-Aware & - & 14.44  & -  & 31.60  & -  & 23.43 & -  & 24.58 & -  & 16.16 & - & 14.78  & - & 20.83 \\
\cellcolor{ctxlevel}5 & \cellcolor{ctxlevel}Ctx-Aware$_\text{tuned}$ & - & 31.60 & -  & 42.55 & -  & 35.44 & -  & 38.44 & -  & 34.14  & - & 28.61  & - & 35.13 \\
\cellcolor{ctxlevel}6 & \cellcolor{ctxlevel}Ctx-Aware$_\text{CPO}$  & - & 33.34 & -  & 43.38 & -  & 37.07 & -  & 39.05 & -  & 34.29 & - & 30.12 & - & 36.21 \\
\cellcolor{sentctxlevel}7 & \cellcolor{sentctxlevel}Sent+Ctx$_{\text{tuned}}$ & 31.91 & 31.88  & 43.57 & 43.62 & 36.06 & 35.96 & \cellcolor{lightred!80}39.77 & 39.62 & 35.58 & 35.75 & 29.19 & 28.83 & 36.01 & 35.94 \\
\cellcolor{sentctxlevel}8 & \cellcolor{sentctxlevel}Sent+Ctx$_{\text{tuned+}}$ & 32.38 & 32.42 & 43.65 & 43.86 & 36.22 & 36.12 & \cellcolor{darkred!80}40.34 & \cellcolor{darkred!80}40.42 & 34.87 & \cellcolor{darkred!80}36.39 & 29.14 & 28.72 & 36.10 & 36.32 \\
\cellcolor{our}9 & \cellcolor{our}CPL (Ours)& 32.28 & 33.26 & \cellcolor{darkred!80}44.22 & \cellcolor{darkred!80}44.24 & \cellcolor{darkred!80}37.80 & \cellcolor{darkred!80}37.78 & 38.80 & \cellcolor{lightred!80}39.73 & \cellcolor{darkred!80}36.51 & 35.75 & 30.03 & 30.06 & 36.61 & \cellcolor{darkred!80}36.80 \\
\cellcolor{our}10 & \cellcolor{our}~~ - IntraLoss & \cellcolor{darkred!80}33.79 & \cellcolor{darkred!80}33.83 & 43.93 & 44.12 & \cellcolor{lightred!80}37.54 & \cellcolor{lightred!80}37.73 & 39.16 & 39.50 & \cellcolor{lightred!80}36.47 & 34.94 & \cellcolor{darkred!80}30.32 & \cellcolor{darkred!80}30.35 & \cellcolor{darkred!80}36.87 & \cellcolor{lightred!80}36.74 \\
\cellcolor{our}11 & \cellcolor{our}~~ - CrossLoss & \cellcolor{lightred!80}33.70 & \cellcolor{lightred!80}33.82 & \cellcolor{lightred!80}44.01 & \cellcolor{lightred!80}44.13 & 37.11 & 37.33 & 39.57 & 39.44 & \cellcolor{lightred!80}36.47 & \cellcolor{lightred!80}35.86 & 29.85 & \cellcolor{lightred!80}30.24 & \cellcolor{lightred!80}36.79 & \cellcolor{darkred!80}36.80 \\

\rowcolor[gray]{0.85}
\Block[c]{1-16}{\texttt{Llama-3-8B-Instruct}}\\
\cellcolor{sentlevel}1 & \cellcolor{sentlevel}Sent-Level & 26.29 & - & 35.30 & -  & 31.16 & -  & 31.59 & -  & 30.37 & -  & 24.33  & - & 29.84 & -  \\
\cellcolor{sentlevel}2 & \cellcolor{sentlevel}Sent-Level$_\text{tuned}$  & 31.80 & - & 42.96 & -  & 35.10 & -  & 38.97 & -  & 38.31 & -  & 27.93  & - & 35.85 & -  \\
\cellcolor{sentlevel}3 & \cellcolor{sentlevel}Sent-Level$_\text{CPO}$ & 33.34 & - & 44.15 & -  & \cellcolor{lightred!80}37.51 & -  & 39.93 & -  & 38.47 & -  & 29.39 & - & 37.13 & -  \\
\cellcolor{ctxlevel}4 & \cellcolor{ctxlevel}Ctx-Aware & - & 13.33  & -  & 18.12  & -  & 15.80 & -  & 16.27 & -  & 13.98 & - & 13.27  & - & 15.13 \\
\cellcolor{ctxlevel}5 & \cellcolor{ctxlevel}Ctx-Aware$_\text{tuned}$ & - & 32.11 & -  & 43.20 & -  & 35.32 & -  & 39.26 & -  & 38.50  & - & 28.14  & - & 36.09 \\
\cellcolor{ctxlevel}6 & \cellcolor{ctxlevel}Ctx-Aware$_\text{CPO}$  & - & 33.49 & -  & 44.13 & -  & 37.67 & -  & 39.42 & -  & 38.46 & - & 29.55 & - & 37.12 \\
\cellcolor{sentctxlevel}7 & \cellcolor{sentctxlevel}Sent+Ctx$_{\text{tuned}}$ & 32.96 & 33.15  & \cellcolor{darkred!80}44.40 & 44.34 & 36.27 & 36.55 & \cellcolor{lightred!80}40.33 & \cellcolor{lightred!80}40.37 & \cellcolor{lightred!80}40.13 & \cellcolor{lightred!80}40.29 & 28.74 & 28.65 & 37.14 & 37.23 \\
\cellcolor{sentctxlevel}8 & \cellcolor{sentctxlevel}Sent+Ctx$_{\text{tuned+}}$ & 32.94 & 33.32 & \cellcolor{lightred!80}44.36 & \cellcolor{darkred!80}44.48 & 36.59 & 36.58 & \cellcolor{darkred!80}40.67 & \cellcolor{darkred!80}40.82 & \cellcolor{darkred!80}40.55 & \cellcolor{darkred!80}40.59 & 28.91 & 28.75 & \cellcolor{lightred!80}37.34 & 37.42 \\
\cellcolor{our}9 & \cellcolor{our}CPL (Ours)& \cellcolor{darkred!80}33.87 & \cellcolor{lightred!80}33.95 & 44.17 & 44.37 & \cellcolor{darkred!80}37.82 & \cellcolor{darkred!80}38.09 & 39.86 & 40.09 & 39.23 & 39.14 & \cellcolor{darkred!80}29.93 & \cellcolor{lightred!80}29.73 & \cellcolor{darkred!80}37.48 & \cellcolor{darkred!80}37.56 \\
\cellcolor{our}10 & \cellcolor{our}~~ - IntraLoss & \cellcolor{lightred!80}33.70 & \cellcolor{darkred!80}34.17 & 44.18 & \cellcolor{lightred!80}44.47 & 37.14 & \cellcolor{lightred!80}37.93 & 39.40 & 39.62 & 39.16 & 39.09 & \cellcolor{lightred!80}29.87 & \cellcolor{darkred!80}30.04 & 37.24 & \cellcolor{lightred!80}37.55 \\
\cellcolor{our}11 & \cellcolor{our}~~ - CrossLoss & 32.90 & 33.75 & 44.09 & 44.05 & 37.24 & 37.43 & 39.44 & 39.52 & 39.09 & 39.01 & 29.48 & 29.34 & 37.04 & 37.18 \\

\bottomrule
\end{NiceTabular}
}
\caption{Results of different systems on sentence-level BLEU metric. Best scores are shown in \colorbox{darkred!80}{dark red} and second-best scores are in \colorbox{lightred!80}{light red}. \textit{Sent} and \textit{CTX} denote sentence-level and context-aware inputs, respectively.}
\label{tab:bleu}
\end{table*}


\section{Experimental Results on X-to-English Translations}\label{apx:en_to_x}

To verify that CPL generalizes beyond the En$\Rightarrow$X setting reported in the main text, we additionally evaluate five X$\Rightarrow$En directions (Es/Fr/Nl/Ru/Zh$\Rightarrow$En) on News Commentary using Qwen3-8B. Tables~\ref{tab:xen_comet} and~\ref{tab:xen_dcomet} report sentence-level COMET and document-level COMET, respectively. Consistent with the main results of En$\Rightarrow$X directions in Table~\ref{tab:comet}, CPL achieves the best scores on all X$\Rightarrow$En directions under both input conditions, and clearly outperforms the joint-training baseline Sent+Ctx$_{\text{tuned}}$ as well as the two ablated variants. The ablations (\#4 vs.\ \#5 vs.\ \#6) again show that both the intra- and cross-condition losses contribute to the final performance.

\begin{table*}[!t]
\centering
\small
\resizebox{\textwidth}{!}{
\begin{NiceTabular}[color-inside]{l|l|rcrcrcrcrc|rc}
\toprule
\Block[l]{2-1}{\textbf{\#}} & \Block[l]{2-1}{\textbf{Model}} & \Block[c]{1-2}{\textbf{Es $\Rightarrow$ En}} & & \Block[c]{1-2}{\textbf{Fr $\Rightarrow$ En}} & & \Block[c]{1-2}{\textbf{Nl $\Rightarrow$ En}} & & \Block[c]{1-2}{\textbf{Ru $\Rightarrow$ En}} & & \Block[c]{1-2}{\textbf{Zh $\Rightarrow$ En}} & & \Block[c]{1-2}{\textbf{\textit{Average}}} & \\
\cmidrule(lr){3-4}\cmidrule(lr){5-6}\cmidrule(lr){7-8}\cmidrule(lr){9-10}\cmidrule(lr){11-12}\cmidrule(lr){13-14}
& & \textit{Sent} & \textit{CTX} & \textit{Sent} & \textit{CTX} & \textit{Sent} & \textit{CTX} & \textit{Sent} & \textit{CTX} & \textit{Sent} & \textit{CTX} & \textit{Sent} & \textit{CTX} \\
\rowcolor[gray]{0.85}
\Block[c]{1-14}{\texttt{Qwen3-8B}}\\
\cellcolor{sentlevel}1 & \cellcolor{sentlevel}Sent-Level$_\text{tuned}$ & 89.13 & - & 87.64 & - & 89.85 & - & 84.76 & - & 86.61 & - & 87.60 & - \\
\cellcolor{ctxlevel}2 & \cellcolor{ctxlevel}Ctx-Aware$_\text{tuned}$ & - & 89.29 & - & 87.79 & - & 90.08 & - & 84.94 & - & 86.85 & - & 87.79 \\
\cellcolor{sentctxlevel}3 & \cellcolor{sentctxlevel}Sent+Ctx$_{\text{tuned}}$ & 89.20 & 89.39 & 87.72 & 87.89 & 90.01 & 90.18 & 84.85 & 85.00 & 86.55 & 86.93 & 87.67 & 87.88 \\
\cellcolor{our}4 & \cellcolor{our}CPL (Ours) & \cellcolor{darkred!80}89.58 & \cellcolor{darkred!80}89.73 & \cellcolor{darkred!80}88.20 & \cellcolor{darkred!80}88.32 & \cellcolor{darkred!80}90.33 & \cellcolor{darkred!80}90.49 & \cellcolor{darkred!80}85.34 & \cellcolor{darkred!80}85.50 & \cellcolor{darkred!80}87.23 & \cellcolor{darkred!80}87.44 & \cellcolor{darkred!80}88.14 & \cellcolor{darkred!80}88.30 \\
\cellcolor{our}5 & \cellcolor{our}~~ - IntraLoss & 89.39 & 89.63 & \cellcolor{lightred!80}88.15 & \cellcolor{lightred!80}88.26 & 90.22 & \cellcolor{lightred!80}90.43 & \cellcolor{lightred!80}85.25 & 85.40 & 87.09 & \cellcolor{lightred!80}87.38 & 88.02 & \cellcolor{lightred!80}88.22 \\
\cellcolor{our}6 & \cellcolor{our}~~ - CrossLoss & \cellcolor{lightred!80}89.50 & \cellcolor{lightred!80}89.66 & 88.05 & 88.22 & \cellcolor{lightred!80}90.24 & 90.40 & 85.21 & \cellcolor{lightred!80}85.43 & \cellcolor{lightred!80}87.14 & 87.37 & \cellcolor{lightred!80}88.03 & 88.22 \\
\bottomrule
\end{NiceTabular}
}
\caption{Results of X-to-English translations on sentence-level COMET using Qwen3-8B. Best scores are shown in \colorbox{darkred!80}{dark red} and second-best scores in \colorbox{lightred!80}{light red}. \textit{Sent} and \textit{CTX} denote sentence-level and context-aware inputs, respectively.}
\label{tab:xen_comet}
\end{table*}

\begin{table*}[!t]
\centering
\small
\resizebox{\textwidth}{!}{
\begin{NiceTabular}[color-inside]{l|l|rcrcrcrcrc|rc}
\toprule
\Block[l]{2-1}{\textbf{\#}} & \Block[l]{2-1}{\textbf{Model}} & \Block[c]{1-2}{\textbf{Es $\Rightarrow$ En}} & & \Block[c]{1-2}{\textbf{Fr $\Rightarrow$ En}} & & \Block[c]{1-2}{\textbf{Nl $\Rightarrow$ En}} & & \Block[c]{1-2}{\textbf{Ru $\Rightarrow$ En}} & & \Block[c]{1-2}{\textbf{Zh $\Rightarrow$ En}} & & \Block[c]{1-2}{\textbf{\textit{Average}}} & \\
\cmidrule(lr){3-4}\cmidrule(lr){5-6}\cmidrule(lr){7-8}\cmidrule(lr){9-10}\cmidrule(lr){11-12}\cmidrule(lr){13-14}
& & \textit{Sent} & \textit{CTX} & \textit{Sent} & \textit{CTX} & \textit{Sent} & \textit{CTX} & \textit{Sent} & \textit{CTX} & \textit{Sent} & \textit{CTX} & \textit{Sent} & \textit{CTX} \\
\rowcolor[gray]{0.85}
\Block[c]{1-14}{\texttt{Qwen3-8B}}\\
\cellcolor{sentlevel}1 & \cellcolor{sentlevel}Sent-Level$_\text{tuned}$ & 87.36 & - & 85.81 & - & 88.43 & - & 82.69 & - & 84.30 & - & 85.72 & - \\
\cellcolor{ctxlevel}2 & \cellcolor{ctxlevel}Ctx-Aware$_\text{tuned}$ & - & 87.66 & - & 86.10 & - & 88.74 & - & 82.98 & - & 84.62 & - & 86.02 \\
\cellcolor{sentctxlevel}3 & \cellcolor{sentctxlevel}Sent+Ctx$_{\text{tuned}}$ & 87.45 & 87.74 & 85.89 & 86.17 & 88.59 & 88.84 & 82.75 & 83.00 & 84.24 & 84.68 & 85.78 & 86.09 \\
\cellcolor{our}4 & \cellcolor{our}CPL (Ours) & \cellcolor{darkred!80}87.84 & \cellcolor{darkred!80}88.10 & \cellcolor{darkred!80}86.41 & \cellcolor{darkred!80}86.62 & \cellcolor{darkred!80}88.91 & \cellcolor{darkred!80}89.13 & \cellcolor{darkred!80}83.31 & \cellcolor{darkred!80}83.57 & \cellcolor{darkred!80}84.96 & \cellcolor{darkred!80}85.27 & \cellcolor{darkred!80}86.29 & \cellcolor{darkred!80}86.54 \\
\cellcolor{our}5 & \cellcolor{our}~~ - IntraLoss & 87.59 & 87.97 & \cellcolor{lightred!80}86.35 & \cellcolor{lightred!80}86.56 & 88.74 & \cellcolor{lightred!80}89.08 & 83.17 & 83.44 & \cellcolor{lightred!80}84.85 & \cellcolor{lightred!80}85.18 & 86.14 & 86.45 \\
\cellcolor{our}6 & \cellcolor{our}~~ - CrossLoss & \cellcolor{lightred!80}87.76 & \cellcolor{lightred!80}88.01 & 86.25 & 86.53 & \cellcolor{lightred!80}88.82 & 89.05 & \cellcolor{lightred!80}83.19 & \cellcolor{lightred!80}83.53 & 84.85 & 85.16 & \cellcolor{lightred!80}86.17 & \cellcolor{lightred!80}86.46 \\
\bottomrule
\end{NiceTabular}
}
\caption{Results of X-to-English translations on document-level COMET using Qwen3-8B. Best scores are shown in \colorbox{darkred!80}{dark red} and second-best scores in \colorbox{lightred!80}{light red}. \textit{Sent} and \textit{CTX} denote sentence-level and context-aware inputs, respectively.}
\label{tab:xen_dcomet}
\end{table*}

\section{Experimental Results on IWSLT2017 Translations}\label{apx:iwslt2017}

We further evaluate CPL on four IWSLT2017 directions (En$\Rightarrow$De/Zh/Fr and Zh$\Rightarrow$En) with Qwen3-8B, where the spoken-language documents exhibit stronger discourse dependencies than News Commentary. Tables~\ref{tab:iwslt_comet} and~\ref{tab:iwslt_dcomet} report sentence-level COMET and document-level COMET, respectively. CPL again yields the best overall performance under both input conditions, with especially large gains on Zh$\Rightarrow$En and En$\Rightarrow$Zh where context is most informative. The only exception is En$\Rightarrow$De under the sentence-level input, where the cross-only variant (\#5) is marginally higher; nevertheless, CPL remains best on average and under the context-aware condition, indicating that combining intra- and cross-condition preferences gives the most robust behavior across directions.

\begin{table*}[!t]
\centering
\small
\resizebox{\textwidth}{!}{
\begin{NiceTabular}[color-inside]{l|l|rcrcrcrc|rc}
\toprule
\Block[l]{2-1}{\textbf{\#}} & \Block[l]{2-1}{\textbf{Model}} & \Block[c]{1-2}{\textbf{En $\Rightarrow$ De}} & & \Block[c]{1-2}{\textbf{En $\Rightarrow$ Zh}} & & \Block[c]{1-2}{\textbf{Zh $\Rightarrow$ En}} & & \Block[c]{1-2}{\textbf{En $\Rightarrow$ Fr}} & & \Block[c]{1-2}{\textbf{\textit{Average}}} & \\
\cmidrule(lr){3-4}\cmidrule(lr){5-6}\cmidrule(lr){7-8}\cmidrule(lr){9-10}\cmidrule(lr){11-12}
& & \textit{Sent} & \textit{CTX} & \textit{Sent} & \textit{CTX} & \textit{Sent} & \textit{CTX} & \textit{Sent} & \textit{CTX} & \textit{Sent} & \textit{CTX} \\
\rowcolor[gray]{0.85}
\Block[c]{1-12}{\texttt{Qwen3-8B}}\\
\cellcolor{sentlevel}1 & \cellcolor{sentlevel}Sent-Level$_\text{tuned}$ & 83.37 & - & 82.67 & - & 82.18 & - & 84.37 & - & 83.15 & - \\
\cellcolor{ctxlevel}2 & \cellcolor{ctxlevel}Ctx-Aware$_\text{tuned}$ & - & 83.65 & - & 83.16 & - & 82.56 & - & 84.52 & - & 83.47 \\
\cellcolor{sentctxlevel}3 & \cellcolor{sentctxlevel}Sent+Ctx$_{\text{tuned}}$ & 83.71 & 83.80 & 82.81 & 83.10 & 82.19 & 82.67 & 84.56 & 84.70 & 83.32 & 83.57 \\
\cellcolor{our}4 & \cellcolor{our}CPL (Ours) & \cellcolor{lightred!80}83.75 & \cellcolor{darkred!80}84.14 & \cellcolor{darkred!80}83.83 & \cellcolor{darkred!80}84.20 & \cellcolor{darkred!80}83.12 & \cellcolor{darkred!80}83.56 & \cellcolor{darkred!80}85.24 & \cellcolor{darkred!80}85.47 & \cellcolor{darkred!80}83.98 & \cellcolor{darkred!80}84.34 \\
\cellcolor{our}5 & \cellcolor{our}~~ - IntraLoss & \cellcolor{darkred!80}83.92 & \cellcolor{lightred!80}84.10 & \cellcolor{lightred!80}83.66 & \cellcolor{lightred!80}84.01 & \cellcolor{lightred!80}82.98 & \cellcolor{lightred!80}83.48 & \cellcolor{lightred!80}85.17 & \cellcolor{lightred!80}85.32 & \cellcolor{lightred!80}83.93 & \cellcolor{lightred!80}84.23 \\
\cellcolor{our}6 & \cellcolor{our}~~ - CrossLoss & 83.74 & 83.95 & 83.59 & 83.98 & 82.92 & 83.43 & 85.11 & 85.29 & 83.84 & 84.16 \\
\bottomrule
\end{NiceTabular}
}
\caption{Results of IWSLT2017 translations on sentence-level COMET using Qwen3-8B (up to 256 context tokens). Best scores are in \colorbox{darkred!80}{dark red} and second-best in \colorbox{lightred!80}{light red}. \textit{Sent} and \textit{CTX} denote sentence-level and context-aware inputs, respectively.}
\label{tab:iwslt_comet}
\end{table*}

\begin{table*}[!t]
\centering
\small
\resizebox{\textwidth}{!}{
\begin{NiceTabular}[color-inside]{l|l|rcrcrcrc|rc}
\toprule
\Block[l]{2-1}{\textbf{\#}} & \Block[l]{2-1}{\textbf{Model}} & \Block[c]{1-2}{\textbf{En $\Rightarrow$ De}} & & \Block[c]{1-2}{\textbf{En $\Rightarrow$ Zh}} & & \Block[c]{1-2}{\textbf{Zh $\Rightarrow$ En}} & & \Block[c]{1-2}{\textbf{En $\Rightarrow$ Fr}} & & \Block[c]{1-2}{\textbf{\textit{Average}}} & \\
\cmidrule(lr){3-4}\cmidrule(lr){5-6}\cmidrule(lr){7-8}\cmidrule(lr){9-10}\cmidrule(lr){11-12}
& & \textit{Sent} & \textit{CTX} & \textit{Sent} & \textit{CTX} & \textit{Sent} & \textit{CTX} & \textit{Sent} & \textit{CTX} & \textit{Sent} & \textit{CTX} \\
\rowcolor[gray]{0.85}
\Block[c]{1-12}{\texttt{Qwen3-8B}}\\
\cellcolor{sentlevel}1 & \cellcolor{sentlevel}Sent-Level$_\text{tuned}$ & 78.27 & - & 76.48 & - & 77.57 & - & 80.45 & - & 78.19 & - \\
\cellcolor{ctxlevel}2 & \cellcolor{ctxlevel}Ctx-Aware$_\text{tuned}$ & - & 78.69 & - & 77.12 & - & 78.23 & - & 80.72 & - & 78.69 \\
\cellcolor{sentctxlevel}3 & \cellcolor{sentctxlevel}Sent+Ctx$_{\text{tuned}}$ & 78.68 & 78.91 & 76.77 & 77.25 & 77.61 & 78.38 & 80.63 & 80.92 & 78.42 & 78.87 \\
\cellcolor{our}4 & \cellcolor{our}CPL (Ours) & \cellcolor{lightred!80}78.95 & \cellcolor{darkred!80}79.39 & \cellcolor{darkred!80}77.64 & \cellcolor{darkred!80}78.19 & \cellcolor{darkred!80}78.62 & \cellcolor{darkred!80}79.22 & \cellcolor{darkred!80}81.45 & \cellcolor{darkred!80}81.73 & \cellcolor{darkred!80}79.17 & \cellcolor{darkred!80}79.63 \\
\cellcolor{our}5 & \cellcolor{our}~~ - IntraLoss & \cellcolor{darkred!80}79.06 & \cellcolor{lightred!80}79.34 & \cellcolor{lightred!80}77.41 & \cellcolor{lightred!80}77.98 & \cellcolor{lightred!80}78.47 & \cellcolor{lightred!80}79.13 & \cellcolor{lightred!80}81.33 & \cellcolor{lightred!80}81.58 & \cellcolor{lightred!80}79.07 & \cellcolor{lightred!80}79.51 \\
\cellcolor{our}6 & \cellcolor{our}~~ - CrossLoss & 78.90 & 79.17 & 77.38 & 77.94 & 78.38 & 79.08 & 81.33 & 81.55 & 79.00 & 79.44 \\
\bottomrule
\end{NiceTabular}
}
\caption{Results of IWSLT2017 translations on document-level COMET using Qwen3-8B (up to 256 context tokens). Best scores are in \colorbox{darkred!80}{dark red} and second-best in \colorbox{lightred!80}{light red}. \textit{Sent} and \textit{CTX} denote sentence-level and context-aware inputs, respectively.}
\label{tab:iwslt_dcomet}
\end{table*}

\section{Effect of Context Length}\label{apx:context_length}
\begin{figure*}[!t]
\centering
\includegraphics[width=0.9\textwidth]{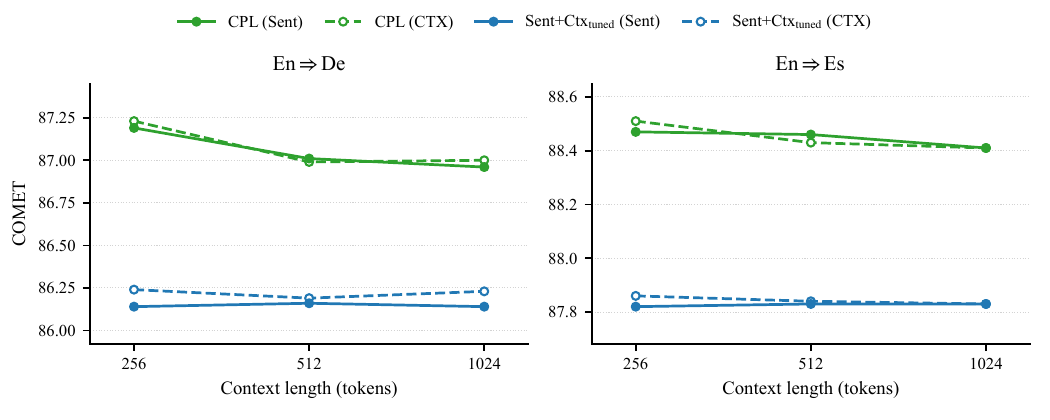}
\caption{COMET under different context lengths (256/512/1024 tokens) on En$\Rightarrow$De and En$\Rightarrow$Es, for sentence-level (\textit{Sent}, solid) and context-aware (\textit{CTX}, dashed) inputs. CPL is shown in green and the Sent+Ctx$_{\text{tuned}}$ baseline in blue.}
\label{fig:context_length}
\end{figure*}

We analyze how the amount of source-side context affects CPL by varying the context length used for context-aware training and inference (256, 512, and 1024 tokens). Figure~\ref{fig:context_length} reports COMET on En$\Rightarrow$De and En$\Rightarrow$Es under both input conditions, together with the joint-training baseline Sent+Ctx$_{\text{tuned}}$. Both systems are largely insensitive to the context length, with COMET varying within a narrow band ($<0.3$) across the three settings. More importantly, CPL consistently stays well above Sent+Ctx$_{\text{tuned}}$ under both conditions and at all context lengths. This indicates that the gains of CPL stem from the preference-based objective itself rather than from a particular context length, and that CPL exploits longer context without being distracted by the additional, potentially noisier input.

\section{Human Evaluation}\label{apx:human_eval}
To complement our automatic and GPT-based evaluation, we conduct two human studies on the En$\Rightarrow$Es task, each using 200 randomly sampled instances scored by annotators who are blind to the system identities.

\noindent\textbf{Study 1: Quality of training preference pairs.} To assess whether our metric-based preference pairs align with human judgments, annotators blind-scored 200 random samples drawn from the CPL training data ($\mathcal{P}_{\text{s}}\cup\mathcal{P}_{\text{c}}\cup\mathcal{P}_{\text{cr}}$). The \textit{chosen} translations scored higher on average than the \textit{rejected} ones (74.97 vs.\ 72.79) and were preferred in 112 cases (tied in 22, rejected in 66). These results indicate that the metric-based preference ordering generally aligns with human judgments, while not ruling out potential metric-induced bias.

\noindent\textbf{Study 2: System-level effectiveness on the test set.} We replicate the GPT-as-Judge analysis in Section~\ref{sec:gpt_as_judge} with human annotators on 200 random test samples. As shown in Table~\ref{tab:human_eval}, and consistent with our GPT-as-Judge findings, CPL outperforms the joint-training baseline under both input conditions, with its context-aware variant achieving the highest score. This indicates that the improvements reported in the paper reflect genuine gains in translation quality as perceived by human annotators.

\begin{table}[!t]
\centering
\small
\begin{tabular}{l|c}
\toprule
\textbf{Model} & \textbf{Avg.\ Score} \\
\midrule
Sent+Ctx$_{\text{tuned}}$ (Sent) & 73.83 \\
Sent+Ctx$_{\text{tuned}}$ (CTX) & 73.44 \\
CPL (Sent) & \cellcolor{lightred!80}73.89 \\
CPL (CTX) & \cellcolor{darkred!80}74.34 \\
\bottomrule
\end{tabular}
\caption{Human evaluation (average total score out of 100) on En$\Rightarrow$Es.}
\label{tab:human_eval}
\end{table}

\section{Distributional Analysis of Context Utility}\label{apx:distribution_analysis}
To better understand how Cross-Preference Learning affects the interaction between sentence-level and context-aware translation, we analyze the distribution of sentences for which one input condition substantially outperforms the other. Specifically, we categorize sentences into five groups based on the COMET score difference between the context-aware and sentence-level outputs, denoted as $\Delta = \text{COMET}_{\text{ctx}} - \text{COMET}_{\text{sent}}$. The categories are defined as follows: \emph{clearly better} ($\Delta \ge 1.0$), \emph{better} ($0.5 \le \Delta < 1.0$), \emph{on par} ($-0.5 < \Delta < 0.5$), \emph{worse} ($-1.0 < \Delta \le -0.5$), and \emph{clearly worse} ($\Delta \le -1.0$).

\begin{figure}[!t]
\centering
\includegraphics[width=0.9\columnwidth, trim={0cm 0.3cm 0cm 0cm}]{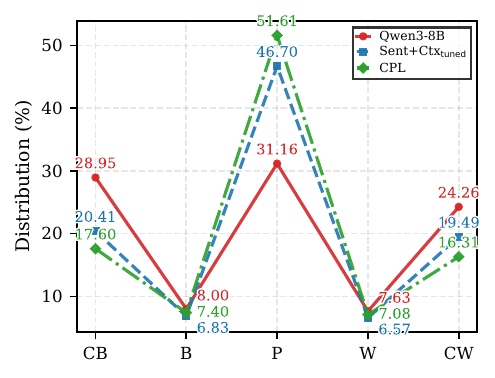}
\caption{Distribution (\%) of sentences by COMET differences between context-aware and sentence-level translations. CB/B/P/W/CW = \emph{Clearly Better}, \emph{Better}, \emph{On Par}, \emph{Worse}, \emph{Clearly Worse}.} 
\label{fig:better_worse}
\end{figure}

As shown in Figure~\ref{fig:better_worse}, the untuned base model, Qwen3-8B, exhibits a large proportion of sentences in the \emph{clearly better, CB} and \emph{clearly worse, CW} categories, indicating that document context can substantially help some sentences while severely degrading others. 

Joint fine-tuning (Sent+Ctx$_{\text{tuned}}$) mitigates this effect by increasing the proportion of \emph{on par, P} cases, suggesting more balanced use of context. CPL further strengthens this trend, achieving the highest share of \emph{on par, P} sentences while simultaneously reducing extreme improvements and degradations. Compared to the Qwen3-8B base model and Sent+Ctx$_{\text{tuned}}$, CPL produces more balanced and stable performance between sentence-level and context-aware translation, reducing extreme gains or degradations associated with context usage. 

\section{Leveraging Both Outputs via Reranking}\label{apx:reranking}
An additional advantage of CPL is that it naturally supports both sentence-level and context-aware translation within a unified framework. Given a source sentence $x$ and its context $c$, CPL produces two candidate translations: a sentence-conditioned output $y$ and a context-conditioned output $y'$. Rather than committing to a single translation mode, these two candidates can be reranked, and the better one selected as the final output.

We report the results in Table~\ref{tab:reranking}. The first two rows show the standalone performance of CPL’s sentence-level and context-aware outputs. We then evaluate three reranking strategies. \textit{Prob.-based Reranking} selects the candidate with higher model likelihood, reflecting the model’s internal confidence. \textit{COMETKiwi-based Reranking} uses a reference-free quality estimator~\cite{rei-2022-cometkiwi}, enabling automatic selection without gold references. \textit{Oracle Reranking} selects the candidate with higher reference-based COMET score and serves as an upper bound. As shown, all reranking strategies consistently improve over individual outputs, demonstrating that sentence-level and context-aware translations provide complementary strengths. Prob.-based reranking yields modest gains, while COMETKiwi achieves larger improvements, suggesting that semantic quality estimation is more effective than likelihood alone.

\begin{table}[!t]
\centering
\small
\begin{tabular}{l|ll}
\toprule
\bf Model & \bf BLEU & \bf COMET \\
\hline
Sent-Level & 36.61 & 87.99\\
Ctx-Aware & 36.80 & 88.06\\
\midrule
\, + Prob. & \cellcolor{lightred!80}37.14 & 88.19\\
\, + COMETKiwi & 37.13 & \cellcolor{lightred!80}88.42\\
\, + Oracle & \cellcolor{darkred!80}38.63 & \cellcolor{darkred!80}88.73\\
\bottomrule
\end{tabular}
\caption{Performance of reranking sentence-level and context-aware translations produced by CPL.}
\label{tab:reranking}
\end{table}

\section{More Analysis Across Input Conditions}\label{apx:calibration}

To examine whether the likelihoods under sentence-level and context-aware inputs are comparable in practice, we measure the normalized log-probability gap between the two input conditions. For each triplet $(x,y,c)$, we compute
\begin{equation}
\delta
= \frac{1}{|y|}\left(
\log p_{\theta}(y \mid x,c)
- \log p_{\theta}(y \mid x)
\right),
\label{equ:calibration_gap}
\end{equation}
where $x$ is the source sentence, $y$ is a target translation, $c$ is the document context, and $\theta$ denotes the trained CPL model. Using En$\Rightarrow$De as an example, we evaluate $\delta$ over all triplets from the preference data, including both $y^{+}$ and $y^{-}$ in each preference pair. The resulting mean and standard deviation are $0.021$ and $0.067$, respectively. The small magnitude and variation of $\delta$ indicate that sentence-level and context-aware inputs induce closely aligned normalized log-probability scales, providing empirical evidence that their likelihoods can be compared in practice.

The per-token normalization in Eq.~\ref{equ:calibration_gap} is used only as a diagnostic to examine the relative probability scales across input conditions. In the CPO and Cross-CPO objectives, we follow CPO~\cite{xu-2024-cpo} and do not apply length normalization. This is unlikely to substantially affect our comparisons because the preferred and dispreferred candidates have very similar lengths. In the News En$\Rightarrow$De cross-condition preference set, the average length difference is only 1.615 words, with a variance of 3.106.

\section{GPT-as-Judge}\label{apx:GPT-as-Judge}
To obtain a more comprehensive and fine-grained quality assessment, we design a detailed evaluation prompt for the judge model. The prompt explicitly specifies the evaluation inputs, scoring dimensions, scoring criteria, and required output format, enabling the model to assess each candidate translation in a more controlled and consistent manner across multiple aspects of quality. The full prompt is shown in Figure~\ref{fig:gpt_evaluation}. 

\section{Case Study}\label{apx:case_study}

Figure~\ref{fig:case_study} presents two examples where CPL produces better translations than Sent+Ctx$_{\text{tuned}}$. In Case 1, both the sentence-level and context-aware variants of Sent+Ctx$_{\text{tuned}}$ incorrectly translate \textit{delusions} as \textit{Delirien}. In contrast, while the sentence-level variant of CPL makes a similar mistake, the context-aware variant correctly translates \textit{delusions} as \textit{Wahnvorstellungen}. This shows that the additional context helps CPL select the more appropriate translation instead of the less suitable \textit{Delirium}. In Case 2, CPL again demonstrates more effective use of context than Sent+Ctx$_{\text{tuned}}$. The sentence-level variant of Sent+Ctx$_{\text{tuned}}$ omits the discourse connective and is less well-formed, while its context-aware variant only partially improves the output by adding \textit{Angesichts dieser Zahlen} but still uses semantically weaker terminology \textit{Verfassungsschutzmaßnahmen} and an imprecise predicate \textit{aufrechterhalten wollen}. In contrast, CPL shows a clear improvement with context: although its sentence-level variant already captures the correct predicate \textit{unterstützen}, the context-aware variant further preserves the full structure of \textit{support retaining} as \textit{die Aufrechterhaltung … unterstützen} and better captures the legal meaning associated with \textit{Roe} as \textit{verfassungsgemäße Schutzgarantie}. This highlights CPL's ability to leverage context for both semantic precision and structural fidelity.

\begin{figure*}[!t]
{\small
\begin{tcolorbox}[title=Case 1, colback=gray!, colframe=black, fonttitle=\bfseries, width=.95\textwidth]
\textbf{[Source]}\\
If a dissident demonstrated against the Soviet system at least twice - say, by distributing illicit writings - the \textcolor{red}{"delusions"} signified schizophrenia.\\
\\
\textbf{[Source-side Context]}\\
Anyone who believed that the political system should be changed was suffering from "reformist delusions." $\dots$ Soviet authorities relied on an idiosyncratic definition of schizophrenia $\dots$\\
\\
\textbf{[Sent+Ctx$_{\text{tuned}}$ (Sent)]}\\
Wenn ein Dissident gegen das sowjetische System mindestens zweimal demonstrierte - etwa indem er illegale Schriften verbreitete - bedeutete das \textcolor{red}{„Delirien“} Schizophrenie.\\
\\
\textbf{[Sent+Ctx$_{\text{tuned}}$ (Ctx)]}\\
Wenn ein Dissident gegen das sowjetische System mindestens zweimal protestiert hatte - etwa durch die Verteilung illegaler Schriften - so wurden die \textcolor{red}{„Delirien“} als Schizophrenie interpretiert.\\
\\
\textbf{[CPL (Sent)]}\\
Wenn ein Dissident mindestens zweimal gegen das sowjetische System demonstriert hatte – etwa indem er illegale Schriften verbreitete –, so bedeutete das \textcolor{red}{„Delirium“}, dass er an Schizophrenie litt.\\
\\
\textbf{[CPL (Ctx)]}\\
Wenn ein Dissident mindestens zweimal gegen das sowjetische System demonstrierte – etwa durch die Verbreitung illegaler Schriften –, dann wurden die \textcolor{red}{„Wahnvorstellungen“} als Zeichen für Schizophrenie interpretiert.\\
\\
\textbf{[Reference]}\\
Wenn sich ein Dissident mindestens zweimal gegen das sowjetische System äußerte, indem er beispielsweise verbotene Schriften in Umlauf brachte, dann deuteten die \textcolor{red}{„Wahnvorstellungen"} auf Schizophrenie hin.
\end{tcolorbox}
\begin{tcolorbox}[title=Case 2, colback=gray!, colframe=black, fonttitle=\bfseries, width=.95\textwidth]
\textbf{[Source]}\\
\textcolor{red}{Given these figures}, it is no wonder that nearly two-thirds of Americans support retaining the constitutional protection that Roe provides.\\
\\
\textbf{[Source-side Context]}\\
Between the Roe decision and 2017 $\cdots$ more than 58 million abortions were performed in the US. $\cdots$ one in four American women will have an abortion $\cdots$\\
\\
\textbf{[Sent+Ctx$_{\text{tuned}}$ (Sent)]}\\
Dass fast zwei Drittel der Amerikaner die Verfassungsschutz durch Roe bewahren wollen, ist kein Wunder.\\
\\
\textbf{[Sent+Ctx$_{\text{tuned}}$ (Ctx)]}\\
Angesichts dieser Zahlen ist es kein Wunder, dass fast zwei Drittel der Amerikaner die \textcolor{red}{Verfassungsschutzmaßnahmen}, die Roe bietet, \textcolor{red}{aufrechterhalten wollen}.\\
\\
\textbf{[CPL (Sent)]}\\
Angesichts dieser Zahlen ist es kein Wunder, dass fast zwei Drittel der Amerikaner die Aufrechterhaltung der Verfassungsschutzgarantie \textcolor{red}{unterstützen}, die Roe bietet.\\
\\
\textbf{[CPL (Ctx)]}\\
Angesichts dieser Zahlen ist es kein Wunder, dass fast zwei Drittel der Amerikanerinnen und Amerikaner die Aufrechterhaltung der verfassungsgemäßen Schutzgarantie, die Roe bietet, unterstützen.\\
\\
\textbf{[Reference]}\\
Angesichts dieser Zahlen verwundert es nicht, dass zwei Drittel aller Amerikanerinnen und Amerikaner sich für den Bestand des verfassungsmäßigen Schutzes aussprechen, der seit Roe besteht.
\end{tcolorbox}
}
\caption{Case studies illustrating that CPL benefits from informative context.}
\label{fig:case_study}
\end{figure*}

\section{Identifying the Usefulness of Context}\label{apx:identifying_context}

To identify test instances for which source-side context is genuinely useful, the prompt shown in Figure~\ref{fig:prompt_dependency_eval} specifically asks whether the current sentence can still be translated accurately and naturally without access to the preceding source context, and whether the context provides necessary disambiguating information. The evaluation covers four dimensions: coreference, lexical ambiguity, discourse/logical relation, and specific referential or background knowledge. Each dimension is scored from 0 to 3, and the total score is mapped to \texttt{Low}, \texttt{Medium}, or \texttt{High} context necessity.

To obtain consistent structured outputs at scale, we use batched API inference with a strict JSON schema (Figure~\ref{fig:dependency_schema}). The schema requires the model to return the total score, the necessity label, a brief explanation of the main translation risk without context, and the most likely error that would arise if the context were ignored.

\begin{figure*}[!t]
\begin{tcolorbox}[title=Prompt Template, colback=gray!5, colframe=black, fonttitle=\bfseries]
\footnotesize

You are a computational linguist with expertise in \texttt{\{tgt\_lang\}}. Your task is to evaluate how strongly the translation of the current sentence depends on its preceding context in a machine translation training example.\\
\\
Your job is \textbf{not} to translate the sentence. Instead, determine whether the \textbf{[current sentence]} can still be translated accurately and naturally into \texttt{\{tgt\_lang\}} \emph{without any context}, and then judge whether the provided \textbf{[context]} supplies \emph{necessary disambiguating information}.\\
\\
Please carefully distinguish the following two cases:\\
(1) The context only makes the translation more natural or complete, but the sentence can still be translated mostly correctly without it $\rightarrow$ \textbf{Low dependency}.\\
(2) Without the context, the sentence would likely be mistranslated due to unresolved ambiguity, incorrect discourse relation, wrong coreference resolution, or incorrect word sense selection $\rightarrow$ \textbf{Medium/High dependency}.\\
\\
Evaluate the sentence on the following four dimensions, assigning a score from 0 to 3 for each:\\
\\
\textbf{1. Coreference Dependency}\\
0: No pronoun/reference, or the reference is clear without context.\\
1: Minor referential uncertainty, but translation is unlikely to be seriously affected.\\
2: Context helps determine the referent; without it, the translation may be awkward or partially wrong.\\
3: Context is necessary to determine referent identity, gender, number, or grammatical role; omission would likely cause mistranslation.\\
\\
\textbf{2. Lexical Ambiguity}\\
0: The key lexical meaning is clear.\\
1: Slight ambiguity, but the most likely translation remains stable.\\
2: Context clearly helps disambiguate the intended sense; without it, the translation may be inaccurate.\\
3: Context is necessary to identify the correct sense; omission would likely cause mistranslation.\\
\\
\textbf{3. Discourse / Logical Relation}\\
0: The sentence is logically self-contained.\\
1: Weak discourse connection, but it does not materially affect translation.\\
2: Context helps determine tone or discourse relation such as contrast, causality, or continuation.\\
3: Without context, the discourse relation would likely be misinterpreted, affecting translation choices.\\
\\
\textbf{4. Specific Referential / Background Knowledge}\\
0: No context-specific referent or background knowledge is needed.\\
1: Slight background dependence, but it does not materially affect translation.\\
2: Context helps identify a specific person, entity, event, or concept.\\
3: Context is necessary to identify the intended referent; omission would likely cause substantial misunderstanding.\\
\\
The total score is the sum of the four dimension scores (0--12).\\
Map the total score to \texttt{necessity\_level} as follows:\\
0--3 $\rightarrow$ \texttt{Low}\\
4--7 $\rightarrow$ \texttt{Medium}\\
8--12 $\rightarrow$ \texttt{High}\\
\\
Output requirements:\\
- In \texttt{bottleneck\_analysis.explanation}, briefly describe the most likely failure point if the context were unavailable.\\
- In \texttt{zero\_shot\_failure\_case}, provide the most likely incorrect interpretation or mistranslation that would arise if the context were completely ignored.\\
- If there is essentially no obvious failure risk, output \texttt{"None"} for \texttt{zero\_shot\_failure\_case}.\\
\\
Return the result strictly following the provided schema, and do not output any additional text.\\
\\
\textbf{[Context]}\\
\texttt{\{context\_text\}}\\
\\
\textbf{[Current sentence]}\\
\texttt{\{src\_sent\}}
\end{tcolorbox}
\caption{Prompt used for context-dependency evaluation of sentence-level translation.}
\label{fig:prompt_dependency_eval}
\end{figure*}

\begin{figure*}[!t]
\begin{tcolorbox}[title=Output JSON Schema, colback=gray!5, colframe=black, fonttitle=\bfseries]
\footnotesize
\begin{verbatim}
{
  "name": "context_dependency_eval_schema",
  "strict": true,
  "schema": {
    "type": "object",
    "additionalProperties": false,
    "properties": {
      "total_dependency_score": {
        "type": "integer",
        "minimum": 0,
        "maximum": 12
      },
      "necessity_level": {
        "type": "string",
        "enum": ["High", "Medium", "Low"]
      },
      "bottleneck_analysis": {
        "type": "object",
        "additionalProperties": false,
        "properties": {
          "has_grammatical_risk": {"type": "boolean"},
          "has_semantic_risk": {"type": "boolean"},
          "explanation": {"type": "string"}
        },
        "required": [
          "has_grammatical_risk",
          "has_semantic_risk",
          "explanation"
        ]
      },
      "zero_shot_failure_case": {"type": "string"}
    },
    "required": [
      "total_dependency_score",
      "necessity_level",
      "bottleneck_analysis",
      "zero_shot_failure_case"
    ]
  }
}
\end{verbatim}
\end{tcolorbox}
\caption{Strict JSON schema used to constrain model outputs in batched API inference.}
\label{fig:dependency_schema}
\end{figure*}

\section{Prompts}
\label{apx:prompts}

In all experiments presented in this paper, we employ a unified set of prompt templates to ensure fair comparison. The two templates share the same instruction style and differ only in whether the source-side context $c$ is provided, so that any performance difference can be attributed to the input condition rather than to prompt design.

As illustrated in Figure~\ref{fig:prompt_sent}, the sentence-level template consists of a system instruction and the user input, where the source sentence $x$ is inserted into the \texttt{\{src\_sent\}} placeholder. In contrast, as shown in Figure~\ref{fig:prompt_ctx}, following \citet{wang-2023-document-level}, the context-aware template augments the input by introducing a \texttt{\{context\_text\}} placeholder for the source-side context $c$ before the source sentence, explicitly instructing the model to incorporate inter-sentence context during translation. In other words, the placeholders \texttt{\{src\_sent\}} and \texttt{\{context\_text\}} correspond to the source sentence $x$ and its context $c$ used throughout the paper, respectively.

Under both templates, the model is required to output only a direct translation of the current source sentence $x$, without any explanation, restatement, or additional content.

\begin{figure*}[!t]
\begin{tcolorbox}[title=Prompt Template, colback=gray!5, colframe=black, fonttitle=\bfseries]
\footnotesize

Translate the text enclosed in triple quotes from \texttt{\{src\_lang\}} into \texttt{\{tgt\_lang\}} without any explanation. \\
\\
\texttt{\{src\_lang\}}: 
''' \texttt{\{src\_sent\}} '''

\end{tcolorbox}
\caption{Prompt used for Sentence-level translation (Sent-Level).}
\label{fig:prompt_sent}
\end{figure*}

\begin{figure*}[!t]
\begin{tcolorbox}[title=Prompt Template, colback=gray!5, colframe=black, fonttitle=\bfseries]
\footnotesize

[\texttt{\{src\_lang\}}]:\texttt{\{context\_text\}} \\
\\
Given the provided context, translate the following \texttt{\{src\_lang\}} sentence to \texttt{\{tgt\_lang\}}.\\
\\
{}[\texttt{\{src\_lang\}}]:\texttt{\{src\_sent\}}
\end{tcolorbox}
\caption{Prompt used for Context-Aware translation (Ctx-Aware).}
\label{fig:prompt_ctx}
\end{figure*}

\section{Training and Inference Cost}\label{apx:inference_efficiency}

Since CPL only modifies the training objective and introduces no architectural change, inference is identical to standard autoregressive decoding and incurs no extra overhead beyond the base model; the two input conditions differ only in input length. All experiments use four NVIDIA L40 GPUs (48~GB each). With LoRA fine-tuning, training any single configuration completes in under 30 minutes, and decoding a full test set with vLLM takes less than 5 minutes under either condition. As the LoRA adapters are merged into the base weights, no extra parameters are stored at inference; for the 8B backbones, the model and its activations fit comfortably within a single L40.

\begin{figure*}[!t]
\begin{tcolorbox}[title=Prompt Template, colback=gray!5, colframe=black, fonttitle=\bfseries]
\footnotesize

You are an expert bilingual translation evaluator.\\
\\
You will receive one JSON object describing a single translation evaluation task. Evaluate 4 candidate translations independently.\\
\\
Input fields:\\
- \texttt{doc\_id}: document id\\
- \texttt{sent\_id}: sentence id\\
- \texttt{lang\_pair}: translation direction, e.g. \texttt{en-de} means translating English into German\\
- \texttt{prompt\_ctx}: broader contextual prompt or discourse context\\
- \texttt{prompt\_sent}: sentence-level translation prompt\\
- \texttt{input}: source sentence\\
- \texttt{ref}: reference translation of the current sentence\\
- \texttt{candidates}: exactly four translations: \texttt{Candidate\_A}, \texttt{Candidate\_B}, \texttt{Candidate\_C}, \texttt{Candidate\_D}\\
\\
Evaluation principles:\\
- Use \texttt{ref} only as a helpful reference for the intended meaning. Multiple valid translations may exist. Do not penalize harmless variation if meaning, contextual fit, and fluency are preserved.\\
\\
For each candidate, consider:\\
1. \texttt{input}\\
2. \texttt{prompt\_sent}\\
3. \texttt{prompt\_ctx}\\
4. \texttt{ref} (Use \texttt{ref} only as a helpful reference for the intended meaning. Multiple valid translations may exist.)\\
5. the candidate itself\\
\\
Score each candidate on 4 dimensions, 25 points each (total 100):\\
\\
\textbf{1. Contextual Appropriateness (25)}\\
Judge whether the translation is fully appropriate given \texttt{prompt\_ctx}, including topic, discourse framing, tone/register, communicative setting, and context-dependent interpretation. Does it correctly resolve ambiguities, pronouns, gender agreements, referential links, or terminology that are dictated by the surrounding context?\\
- 22--25: fully context-appropriate\\
- 16--21: mostly appropriate, with minor contextual mismatch\\
- 0--15: contextually inappropriate, inconsistent, or fails to resolve context-dependent meaning correctly\\
\\
\textbf{2. Faithfulness and Accuracy (25)}\\
Judge whether the candidate accurately conveys the meaning of \texttt{input}, using \texttt{ref} as a strong anchor.\\
- 22--25: highly accurate\\
- 16--21: mostly accurate, minor deviations\\
- 0--15: major semantic errors, omissions, additions, or distortions\\
\\
\textbf{3. Fluency and Naturalness (25)}\\
Judge grammaticality, idiomaticity, readability, and naturalness in the target language.\\
- 22--25: fluent and natural\\
- 16--21: understandable but somewhat awkward\\
- 0--15: noticeable language problems\\
\\
\textbf{4. Terminology and Expression Quality (25)}\\
Judge lexical choice, phrasing, precision, and stylistic suitability, especially for key expressions.\\
- 22--25: precise and well-phrased\\
- 16--21: acceptable but somewhat suboptimal\\
- 0--15: weak, awkward, or inappropriate wording\\
\\
Output requirements:\\
- Score each candidate and provide one concise overall reasoning.\\
- In contextual reasoning, explicitly mention how \texttt{prompt\_ctx} affects the judgment.\\
- Ensure \texttt{total} equals the sum of the four dimension scores.\\
- Return the result strictly following the provided JSON schema without extra text.
\end{tcolorbox}
\caption{Prompt used for GPT-as-Judge.}
\label{fig:gpt_evaluation}
\end{figure*}

\end{document}